\documentclass[sigconf]{aamas}

\usepackage{balance} 
\usepackage{enumitem}
\usepackage{graphicx}
\usepackage{subfig}
\usepackage{wrapfig}

\usepackage{tcolorbox}
\usepackage{listings}
\usepackage{xcolor}
\usepackage[utf8]{inputenc}

\lstdefinestyle{mystyle}{
    basicstyle=\ttfamily\small,        
    breaklines=true,                   
    columns=flexible,                  
    frame=none,                        
    backgroundcolor=\color{gray!5},    
    rulecolor=\color{gray!75},         
    numbers=none,                      
    breakatwhitespace=true,             
    breakindent=0pt                     
}

\setcopyright{ifaamas}
\acmConference[AAMAS '25]{Proc.\@ of the 24th International Conference
on Autonomous Agents and Multiagent Systems (AAMAS 2025)}{May 19 -- 23, 2025}
{Detroit, Michigan, USA}{A.~El~Fallah~Seghrouchni, Y.~Vorobeychik, S.~Das, A.~Nowe (eds.)}
\copyrightyear{2025}
\acmYear{2025}
\acmDOI{}
\acmPrice{}
\acmISBN{}

\acmSubmissionID{<<67>>}

\title[AAMAS-2025 Formatting Instructions]{Online Preference-based Reinforcement Learning with Self-augmented Feedback from Large Language Model}

\author{Songjun Tu}
\affiliation{
  \institution{Institute of Automation, CASIA}
  \institution{Peng Cheng Laboratory}
  \institution{School of Artificial Intelligence, UCAS}
  \country{China}}
\email{tusongjun2023@ia.ac.cn}

\author{Jingbo Sun}
\affiliation{
  \institution{Institute of Automation, CASIA}
  \institution{Peng Cheng Laboratory}
  \institution{School of Artificial Intelligence, UCAS}
  \country{China}}
\email{sunjingbo2022@ia.ac.cn}

\author{Qichao Zhang}
\affiliation{
  \institution{Institute of Automation, CASIA}
  \institution{School of Artificial Intelligence, UCAS}
  \country{China}}
\email{zhangqichao2014@ia.ac.cn}

\author{Xiangyuan Lan}
\affiliation{
  \institution{Peng Cheng Laboratory}
  \country{China}}
\email{lanxy@pcl.ac.cn}

\author{Dongbin Zhao}
\affiliation{
  \institution{Institute of Automation, CASIA}
  \institution{Peng Cheng Laboratory}
  \institution{School of Artificial Intelligence, UCAS}
  \country{China}}
\email{dongbin.zhao@ia.ac.cn}

\begin{abstract}
Preference-based reinforcement learning (PbRL) provides a powerful paradigm to avoid meticulous reward engineering by learning rewards based on human preferences. 
However, real-time human feedback is hard to obtain in online tasks. 
Most work suppose there is a "scripted teacher" that utilizes privileged predefined reward to provide preference feedback. 
In this paper, we propose a RL Self-augmented Large Language Model Feedback (RL-SaLLM-F) technique that does not rely on privileged information for online PbRL. 
RL-SaLLM-F leverages the reflective and discriminative capabilities of LLM to generate self-augmented trajectories and provide preference labels for reward learning.
First, we identify a failure issue in LLM-based preference discrimination, specifically "query ambiguity", in online PbRL. Then LLM is employed to provide preference labels and generate self-augmented imagined trajectories that better achieve the task goal, thereby enhancing the quality and  efficiency of feedback. 
Additionally, a double-check mechanism is introduced to mitigate randomness in the preference labels, improving the reliability of LLM feedback.
The experiment across multiple tasks in the MetaWorld benchmark demonstrates the specific contributions of each proposed module in RL-SaLLM-F, and shows that self-augmented LLM feedback can effectively replace the impractical "scripted teacher" feedback.
In summary, RL-SaLLM-F introduces a new direction of feedback acquisition in online PbRL that does not rely on any online privileged information, offering an efficient and lightweight solution with LLM-driven feedback.\footnote{Corresponding author: Qichao Zhang, zhangqichao2014@ia.ac.cn \\
Code Page: \href{https://github.com/TU2021/RL-SaLLM-F}{https://github.com/TU2021/RL-SaLLM-F}}

\end{abstract}

\keywords{Online Preference-based Reinforcement Learning, Self-augmented LLM, LLM-driven Feedback, Query Ambiguity.}

\newcommand{\BibTeX}{\rm B\kern-.05em{\sc i\kern-.025em b}\kern-.08em\TeX}

\begin{document}


\pagestyle{fancy}
\fancyhead{}


\maketitle 

\section{Introduction}
Designing complex artificial reward functions is labor-intensive and time-consuming for reinforcement learning (RL) \cite{irlsurvey21,gupta2022unpacking}.
Preference-based RL (PbRL) is considered a key paradigm to address this challenge by learning
rewards based on human preference \cite{christiano2017deep,pebble21}. 

However, the substantial human effort required to label a large number of preference queries significantly hinders its widespread application in real-world scenarios \cite{wirth2017survey,liang2022reward}.
Especially for online PbRL \cite{park2022surf}, obtaining real-time preference feedback necessitates immediate human interaction with the environment.
Current online PbRL methods often assume a "scripted teacher" that provides real-time preference feedback by comparing handcrafted rewards of two trajectories from replay buffer \cite{lee2021b,pebble21}. 
Unfortunately, relying on privileged rewards undermines the original intent of PbRL.

Large Pre-trained Models (LPMs), such as large language models (LLMs) \cite{zhao2023survey} and vision-language models (VLMs) \cite{zhang2024vision}, equipped with extensive human prior knowledge, have recently gained significant attention. 
Recently, some studies have explored using LPMs instead of human supervision for reward design, including generating reward code \cite{eureka24,robogen24,zeng2024learning} or calculating dense rewards directly based on the comparison of policy trajectories and targets \cite{ma2023liv,rocamonde2024vision,sontakke2024roboclip}. 
Unfortunately, the first type of approach requires access to the simulation environment's code, and evaluating the reward code often involves multiple full RL training cycles, which is impractical for real-world applications.
The other type of approach relies on comparing image-text similarities, but a single image may fail to reflect the underlying dynamic information described in the text, and reward variance is easily introduced by visual noise in the images \cite{wang2024rl}.
Instead of designing reward functions or codes, PbRL learns rewards by comparing trajectory pairs. This approach requires only a single complete online training cycle and does not necessitate access to any low-level information, such as the environment's code \cite{wang2024rl,pebble21}.   
The learned downstream policies are guaranteed to have a suboptimal performance bound \cite{zhu2023principled}.

Instead of relying on "scripted teacher", some works \cite{wang2024rl,wang2024prefclm} try to obtain preference labels by querying LPMs with manually designed prompts, to train rewards and policies that align with human intentions.
Although LPMs have substantial capabilities in analyzing trajectories, we find that they still struggle to distinguish the quality of suboptimal trajectories generated by the poor policy, especially during the early stages of training.
The failure in preference discrimination impacts the learning of the reward model, further hindering performance.
We refer to this phenomenon as "query ambiguity."
Recently, LPMs have demonstrated reflective \cite{yuanself24, mcaleese2024llm, zheng2024planagent} and planning abilities in decision-making tasks, understanding the environment and predicting or planning future actions to achieve the task goal \cite{zheng2024preliminary,zhai2024fine,park2023generative,chen2023robogpt,chen2024common}. 
These works inspire us with the following thought:

\textbf{\textit{In addition to the discriminative capability, can LPMs leverage reflection to generate self-augmented trajectories that promoting efficient reward learning in online PbRL?}}

In this paper, we propose {\textbf{R}einforcement \textbf{L}earning from \textbf{S}elf-\textbf{a}ugmented \textbf{LLM} \textbf{F}eedback (RL-SaLLM-F). }
This method seeks to mitigate query ambiguity in online PbRL and replaces "scripted teacher" with LLM-driven feedback.
RL-SaLLM-F operates without relying on any predefined privileged rewards, thereby establishing a practical new paradigm for online PbRL.
Specifically, state trajectories in the replay buffer are converted into text descriptions, and LLMs are queried to assign preference labels based on these trajectories. 
Furthermore, a second round of queries is performed to prompt the LLM to generate imagined trajectories aligned with task goals, serving as as new preference pairs for reward learning. 
In addition, we mitigate the randomness of LLM-based preference labels using a double-check mechanism that swaps the order of two input trajectories.
Finally, we evaluate RL-SaLLM-F on multiple tasks in the Metaworld  benchmark \cite{yu2020meta}, it achieves comparable or better success rates than  feedback from "scripted teacher" with privileged rewards.
The core contributions in this paper are as follows:
\begin{itemize}
\item We identify an issue of query ambiguity in online PbRL with LLM feedback, and propose the RL-SaLLM-F method to mitigate the potential challenges.
\item We leverage self-augmented LLM feedback to obtain preference labels efficiently and use a double-check mechanism to reduce randomness in the LLM-based labels. 
\item RL-SaLLM-F achieves comparable performance to that of "scripted teacher" with privileged reward information in the Metaworld benchmark, requiring only the lightweight and cost-effective GPT-4o-mini.
\item The overall framework of RL-SaLLM-F does not rely on any predefined rewards or real-time human interaction, establishing a practical new paradigm for online PbRL.
\end{itemize}
\section{Related Works}

\begin{figure*}[t]
\centering
\subfloat[The sampled trajectory pairs that the LLM failed to assess.]{
\includegraphics[width=.47\linewidth]{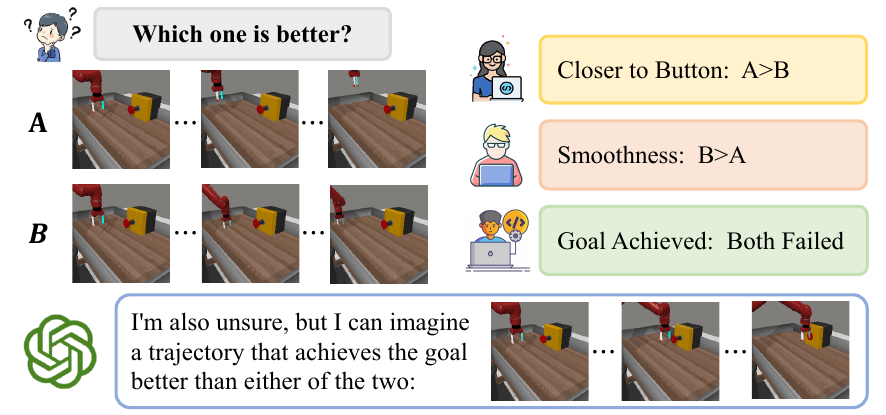}
\label{fig:motivation1}
}
\subfloat[Training curves of LLM feedback.]{
\includegraphics[width=.26\linewidth]{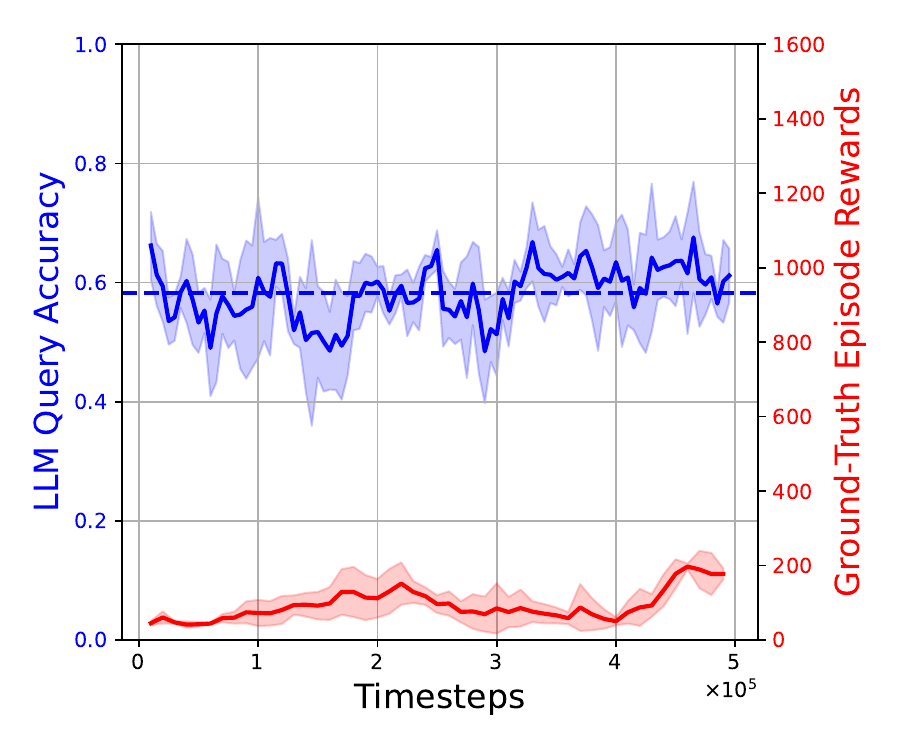}
\label{fig:motivation2}
}
\subfloat[Double-check \& Extra self-augmentation.]{
\includegraphics[width=.26\linewidth]{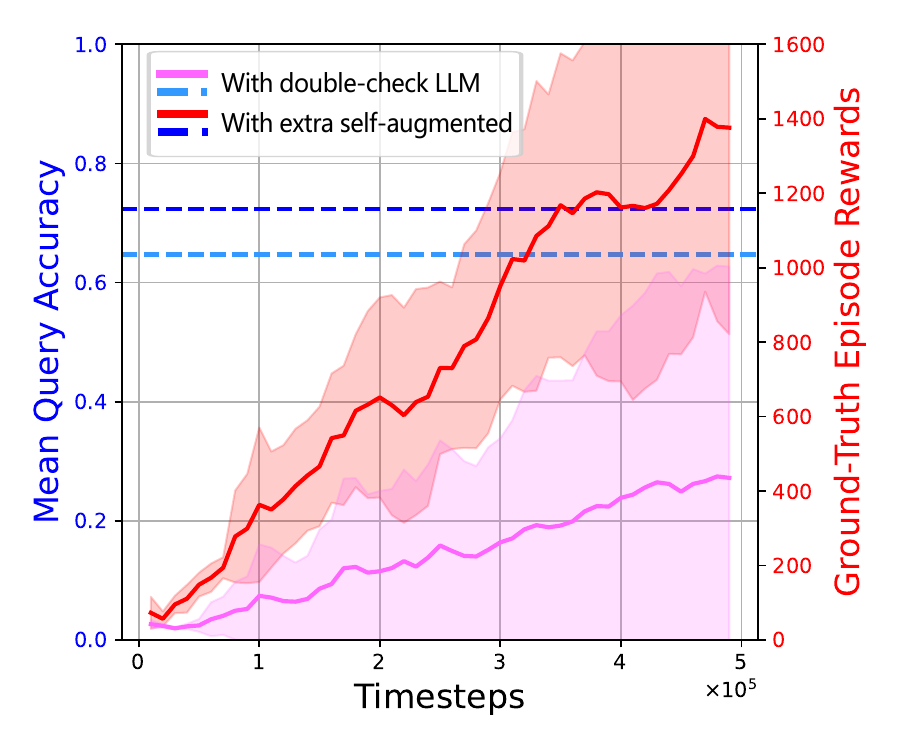}
\label{fig:motivation3}
}
\caption{
    Query ambiguities in online PbRL. 
    (a) Two failed trajectories are provided, converted into text form, and input into the LLM, showing that the LLM struggles to evaluate such trajectories. Specific examples can be found in Appendix \ref{app:d}; 
    (b) Training curves of PEBBLE with LLM feedback. The blue line represents the labeling accuracy of the LLM and the red line represents the predefined episode rewards.
    (c) Training curves of double-check mechanism and additional self-augmented LLM feedback.
}
\label{fig:motivation}
\vspace{-0.2cm}
\end{figure*}

\subsection{Large Pre-trained Models as Rewards}
Applying RL in reward-free environments is challenging.
Some studies assist in reward code design with the perceptual capabilities of LPMs \cite{eureka24, text2reward24,robogen24,zeng2024learning,codeasr24}.
A notable example is Eureka \cite{eureka24}, which leverages GPT-4 \cite{achiam2023gpt} to evaluate environment and task information and generate reward code, followed by iterative updates using evolutionary algorithms.
Building on this, \cite{zeng2024learning} introduces a reflective mechanism to achieve further self-alignment.
However, these methods assume that the environment code is accessible, and each evaluation of reward code requires a full RL training cycle, which is impractical for real-world deployment.
Instead of designing reward code directly, we utilizes a LLM to provide preference labels for trajectory pairs from online replay buffer, relying on a more accessible textual description of the environment rather than full access to its code.

Another line of research obtains rewards from visual observations by VLMs \cite{mahmoudieh2022zero, ma2023liv, rocamonde2024vision,sontakke2024roboclip,luo2024text}. 
For example, RoboCLIP \cite{sontakke2024roboclip} computes reward signals by comparing video representations of expert demonstrations with those of policy trajectories.
Recently, FuRL \cite{fu2024furl} incorporates a learnable layer into CILP, and fine-tunes it to align with real tasks.
Despite achieving impressive performances, these methods often need to fine-tune with expert data to reduce variance and noise in the reward  \cite{fu2024furl,sontakke2024roboclip}.
On the other hand, some works compare image-text similarities to obtain dense rewards, {but a single image is insufficient to capture the required dynamic information \cite{wang2024rl}}. 
For example, in robotic tasks like picking up an object, a single image cannot indicate if the arm is approaching or moving away, causing confusion reward learning.
Different from the aforementioned methods, we abstracts observed trajectories into text and learns a reward model through pairwise trajectory comparisons.
Our approach enhances the ability to understand trajectories, reduces query costs, and guarantees the convergence properties of downstream RL with preference-based reward modeling \cite{zhu2023principled}.

\subsection{Preference-based Reinforcement Learning}
A new paradigm for learning reward functions during interactions with the environment, known as online PbRL, leverages human feedback given in the form of trajectory preferences \cite{christiano2017deep,pebble21}.  
Most of online PbRL research aims to address challenges such as preference noise \cite{tu2024dataset}, reward credit assignment \cite{prior24}, limited preference data \cite{qpa24} and efficient querying \cite{park2022surf}. However, frequent querying of human preferences during online training is impractical. In previous online PbRL methods, a "scripted teacher" is assumed to provide preference labels by comparing privileged predefined rewards of two trajectories in the specific task \cite{lee2021b}.

This "scripted teacher" relying on privileged rewards serves as a compromise to investigate online PbRL algorithms as immediate human feedback is not available.
In this work, we aim to replace the "scripted teacher" with a LLM to provide preference feedback. 
Recently, existing works such as RL-VLM-F \cite{wang2024rl} and PrefCLM \cite{wang2024prefclm} have explored using VLMs to evaluate the quality of trajectories for visual input tasks.
However, repeated image pair queries \cite{wang2024rl} and the aggregation of multiple GPT4 feedback \cite{wang2024prefclm} make these methods costly.  
Additionally, we find samples with no significant preference difference in the replay buffer can lead to a potential risk of query ambiguity, which will further increase costs and reduce efficiency.
In contrast, we use a self-augmented and double-checking feedback to improve the efficiency and reliability based on two-round preference queries, mastering robotic manipulation tasks with with cost-effective GPT-4o-mini \cite{achiam2023gpt}.


\section{Preliminaries}
\subsection{Reinforcement Learning}
RL is formulated as a Markov Decision Process (MDP) \cite{rl18}, which is characterized by the tuple $M=\langle S, A, P, r,\gamma \rangle$.  $S$ is the state space, $A$ is the action space, $P: S \times A \times S \rightarrow \mathbb{R}$ is the transition probability distribution, $r: S \rightarrow \mathbb{R}$ is the reward function, and $\gamma \in (0, 1)$ is the discount factor.
The objective of RL is to determine an optimal policy $\pi^{}$ that maximizes the expected cumulative reward: $
\pi=\arg \max_\pi \mathbb{E}_{s_0, a_0, \ldots}\left[\sum_{t=0}^{\infty} \gamma^t r\left(s_t\right)\right]$.

We use Soft Actor-Critic (SAC) \cite{sac18}, an off-policy RL algorithm, as our low-level approach. 
Specifically, transitions from interactions with the environment are stored in the replay buffer. The actor-critic is then trained by sampling data from the buffer and maximizing the entropy of the stochastic policy.

\subsection{Learning Rewards from Human Feedback}

Following previous studies \cite{aligndiff24}, we consider state-only trajectories of length $H$  composed of states and actions, defined as $\sigma = \{s_k, \ldots, s_{k+H}\}$. 
The goal is to align human preference $y$ between pairs of trajectory segments $\sigma^0$ and $\sigma^1$, where $y$ denotes a distribution indicating human preference, captured as $y \in \{1, 0, 0.5\}$. 
The preference label $y=1$ indicates that $\sigma^0$ is preferred to $\sigma^1$, namely, $\sigma^0 \succ \sigma^1$, $y=0$ indicates $\sigma^1 \succ \sigma^0$, and $y=0.5$ indicates equal preference for both. 
The preference data are stored as triples, denoted as $\mathcal{D}$: $(\sigma^0, \sigma^1, y)$.
Then, the Bradley-Terry model \cite{bt95} is employed to couple preferences with rewards. The preference predictor is defined as follows: 
\begin{equation}
    P_{\psi}[\sigma^1 \succ \sigma^0] = \frac{\exp\left(\sum_t \hat{r}_\psi(s_t^1)\right)}{\sum_{i \in \{0,1\}} \exp\left(\sum_t \hat{r}_\psi(s_t^i)\right)}
\end{equation}
where $\hat{r}_\psi$ is the reward model to be trained, and $\psi$ is its parameters.
Subsequently, the reward function is optimized using the cross-entropy loss, incorporating the human predefined label $y$ and the preference predictor $P_{\psi}$:
\begin{equation}
\begin{aligned}
    \mathcal{L}_{\text{CE}} = -\mathbb{E}_{(\sigma^0,\sigma^1,y) \sim \mathcal{D}}  & \Big\{ (1-y) \log P_{\psi}[\sigma^0 \succ \sigma^1] \\
    + & y \log P_{\psi}[\sigma^1 \succ \sigma^0] \Big\}
\label{eqt:pref_ce}
\end{aligned}
\end{equation}

In online PbRL, we train an off-policy RL algorithm while periodically sampling trajectory pairs from the replay buffer $\mathcal{B}$ for preference queries, as done in PEBBLE \cite{pebble21}.
As online human preference is often not available, the "scripted teacher" is used to provide
preference labels $y$ in most previous works \cite{christiano2017deep,pebble21}. 
Specifically, it generates preferences based on predefined task reward $r$ from the environment as follows: $y=i$, where $i=\arg \max_{i \in \{0,1\}} \sum_{t} r(s_t^i,a_t^i)$.
These trajectory pairs, along with the preference labels, form the preference dataset $\mathcal{D}$. 
We train the reward model on $\mathcal{D}$, and subsequently relabel the data in 
$\mathcal{B}$ using this model.

\begin{figure*}[htb]
  \centering
  \includegraphics[width=\linewidth]{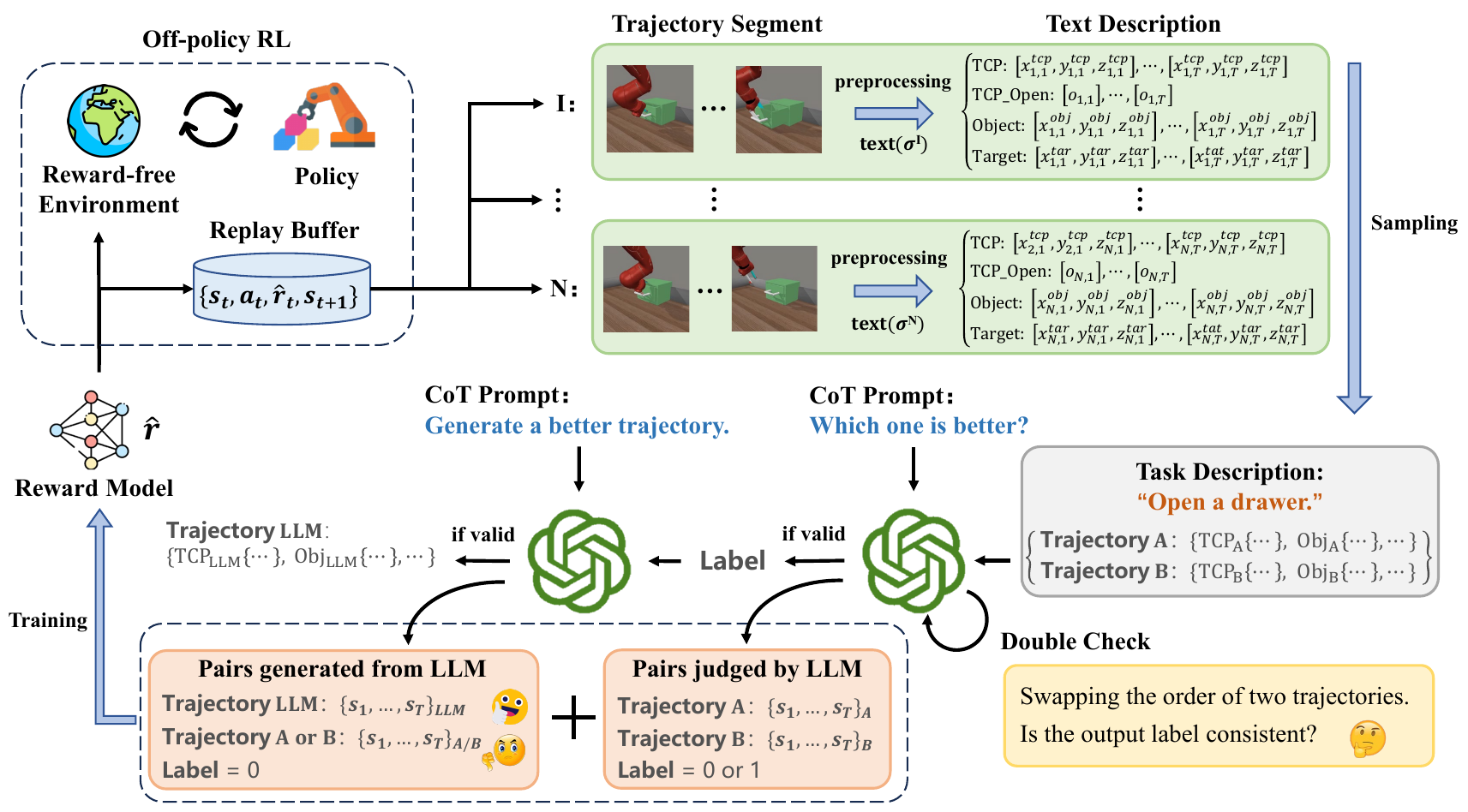}
  \caption{The overall framework of RL-SaLLM-F. 
    First, trajectories are sampled from the replay buffer and converted into coordinate text descriptions. 
    Next, the text representation of the trajectory pairs are selected and queried through the LLM twice with different orderings to obtain feedback labels. 
    Subsequently, based on the sampled trajectories, the 'imagined' trajectories that better achieve the goal are generated by the LLM to train the reward model.
    }
  \label{fig:main}
  \vspace{-0.2cm}
\end{figure*}

\section{Query Ambiguity in online PbRL}
\label{sec4}
In this section, we highlight a potential risk of query ambiguity in online PbRL.
This risk stems from the inherent challenges due to the trajectories sampled from the replay buffer tend to be suboptimal and of insufficient quality.
Figure \ref{fig:motivation1} illustrates an example from the MetaWorld Button-Press task, with two suboptimal trajectories sampled from the replay buffer.
Neither trajectory successfully achieves the target: trajectory A deviates upward with larger movement amplitude, while trajectory B moves to the left-rear with smaller movement amplitude.
For such low-quality trajectories, human may be able to evaluate their quality from multiple perspectives (e.g., distance to the button, trajectory smoothness), but it is challenging to determine which trajectory is better in terms of achieving the target.
The "scripted teacher" obtains labels from privileged task rewards by comparing the rewards of two trajectories and selecting the one with the higher cumulative rewards as the preferred trajectory, thereby avoiding this potential issue.
However, when human or LLM provide feedback evaluations, the situation becomes significantly more complex.
In Figure \ref{fig:motivation2}, we train  PEBBLE \cite{pebble21}  by directly replacing the "scripted teacher" with GPT-4o-mini as feedback, and record the label accuracy (compared to the ”scripted teacher“) and episode rewards.
The policy performance remain poor throughout the training, and the accuracy of the LLM labels is consistently low, with an average accuracy of only 58.2\%.

How can we improve policy learning in online PbRL? 
An intuitive approach is to enhance the reliability of the labels. 
In Section \ref{sec5.2}, we reduce the randomness of the labels through a double-check mechanism, which further improve the accuracy of the LLM queries.
In Figure \ref{fig:motivation3}, we plot the training curves and average label accuracy (pink and cyan lines) after incorporating the double-check mechanism. 
The label accuracy increase from 58.2\% to 64.8\%, which accompanies an improvement in policy performance. 
However, it still demonstrates low sample efficiency and fails to achieve the task goal.

Another idea is, can we use additional high-quality data to train the reward model, thereby driving both reward and policy learning?
In Section \ref{sec5.3}, we propose a self-augmented LLM feedback method, which generates high-quality additional trajectory pairs to facilitate the learning of the reward model. 
Similarly, we plot the corresponding training curves and average accuracy (red and blue lines) in Figure \ref{fig:motivation3}.
Note that we do not enhance the discriminative ability of LLM by prompt engineering; instead, we simply use the higher-quality data generated by the LLM to train the reward model.
As the training policy gradually approaches optimality, the average label accuracy further increases from 64.8\% to 72.3\%.

Based on these observations, we think self-augmented LLM feedback is effective for mitigating the risk of query ambiguity in online PbRL. 
This approach not only directly improves the discriminative reliability of the LLM but also further drives reward learning through self-augmented data.
In turn, the highly discriminative reward function encourages the policy to sample diverse, high-quality data, enriching the LLM's query diversity and indirectly enhancing its discriminative accuracy, forming a positive feedback loop in the training process.


\section{RL-SaLLM-F}
\label{sec5}

\subsection{The Overall Framework}
In this section, we propose \textbf{R}einforcement \textbf{L}earning from \textbf{S}elf-\textbf{a}ugmented \textbf{LLM} \textbf{F}eedback (RL-SaLLM-F).

The overall training process is shown in Figure \ref{fig:main}, following PEBBLE\cite{pebble21}, with an off-policy SAC agent
as the policy, consisting of three steps:

\noindent
\begin{itemize}[leftmargin=*]
    \item \textbf{Step 1 (Unsupervised Pre-training)}:
    In the early stage of training, to encourage exploration and increase trajectory diversity, an intrinsic reward $r^{\text{int}}(\mathbf{s}_t) = \log(\|\mathbf{s}_t - \mathbf{s}_t^k\|)$ is used for pre-training downstream policy $\pi_\phi$, here $s_t^k$ is the $k$-th nearest neighbor of $s_t$.
    \item \textbf{Step 2 (Label Querying and Reward Learning)}: 
    Sampling candidate trajectory pairs and querying labels from the LLM (see Section \ref{sec5.2}), followed by generating self-augmented imagined trajectory to train the reward model $\hat{r}_{\psi}$ (see Section \ref{sec5.3}).
    \item \textbf{Step 3 (Policy Learning)}: Relabeling the replay buffer $\mathcal{B}$ with $\hat{r}_\psi$ and updating the policy $\pi_\phi$ and Q-function $Q_\theta$ using SAC.
    \item \textbf{Repeat Step 2 and Step 3}.
\end{itemize}

\subsection{Query Feedback Labels Judged by LLM}
\label{sec5.2}
First, we sample a trajectory pair $\{ \sigma^0,\sigma^1 \}$ from the replay buffer $\mathcal{B}$, convert it into textual representations $\{ \text{text}(\sigma^0), \text{text}(\sigma^1) \}$, and query the LLM with task-directed Chain-of-Thought (CoT) prompt  for preference labels:
\begin{equation}
    y = \text{LLM}\left(\text{Task}, \text{Traj 0}=\text{text}(\sigma^0), \text{Traj 1}=\text{text}(\sigma^1) \right)
\end{equation}

To improve query accuracy, we use a double-check mechanism to  mitigate the randomness in the labels. Specifically, we reverse the order of the input trajectories and query the LLM again:
\begin{equation}
    y_{inv} = \text{LLM}\left(\text{Task}, \text{Traj 0}=\text{text}(\sigma^1), \text{Traj 1}=\text{text}(\sigma^0) \right)
\end{equation}

If the values of $y$ and $y_{inv}$ indicate the same preference feedback is labeled, namely $y = 0, y_{inv} = 1$ or $y = 1, y_{inv} = 0$, we consider the feedback label to be valid. 
Otherwise, we consider this trajectory pair to be indistinguishable for the LLM, and discard it.
When the label is valid, the trajectory pair triple $(\sigma^0, \sigma^1, y)$ is stored into the preference dataset $\mathcal{D}$.

\textbf{Remark 1.} 
\textit{
An alternative is to consider the LLM's evaluation of the trajectory pair as equivalent, retaining the pair and setting the label as $y=0.5$, treating it as soft label in reward  learning. 
We compare this approach and find that, during early training, the LLM struggle to distinguish the quality of sampled trajectories, resulting in numerous hallucinated labels. 
Including such trajectory pairs is detrimental to reward model learning, so we conclude that discarding them leads to more efficient reward training.
}

\begin{figure*}[t]
  \centering
  \includegraphics[width=\linewidth]{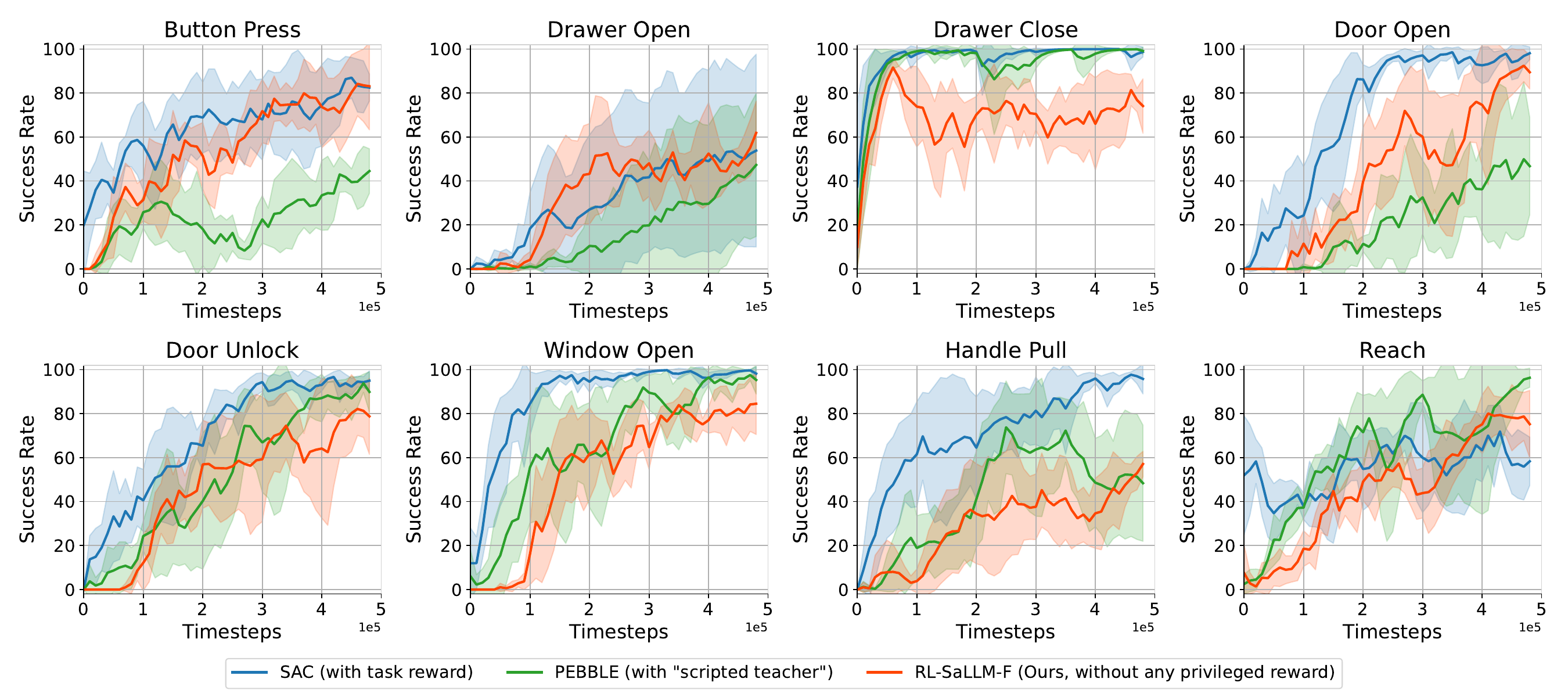}
  \caption{Learning curves of all compared methods on 8 tasks. 
  Results are averaged over 5 seeds, and shaded regions represent standard error.
  RL-SaLLM-F masters robotic manipulation without any online privileged reward, performing on par with PEBBLE, which uses 'scripted teacher' feedback, and even SAC with predefined reward functions in partial tasks.
  }
  \label{fig:exp_main}
\end{figure*}

\begin{figure*}[t]
  \centering
  \includegraphics[width=\linewidth]{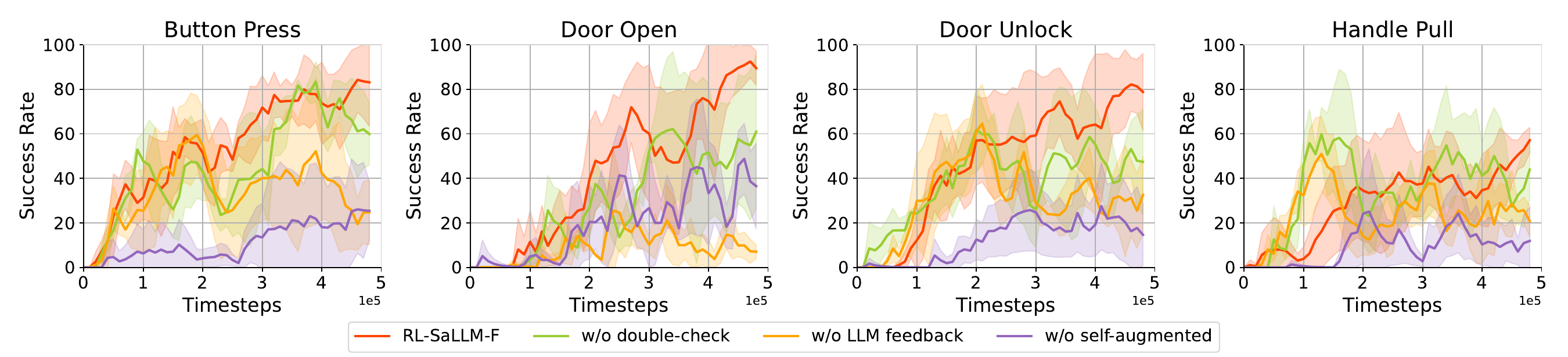}
  \caption{Learning curves of the ablation study. When any component of RL-SaLLM-F is removed, the performance decreases. 
  Specifically, the absence of self-augmented feedback leads to notably poor success rate.
  }
  \label{fig:abl}
\end{figure*}

\subsection{Self-augmented Feedback Generated by LLM}
\label{sec5.3}
Since the differences between trajectory pairs sampled from the replay buffer are small, particularly during early training, we leverage the reflective and planning abilities of LLM to generate an imagined trajectory that is more goal-directed to accelerate reward model training. Specifically, we prompt the LLM to generate a textual trajectory $\text{text}({\sigma^{LLM}})$ that outperform the better trajectory $\sigma^{0/1}$ of the current trajectory pair $\{ \sigma^0,\sigma^1 \}$, while ensuring that it shares the same initial state as the input:
\begin{equation}
    \text{text}(\sigma^{LLM}) = \text{LLM}\left(\text{Generate Prompt}, \text{Traj}=\text{text}(\sigma^{0/1})  \right)
\end{equation}

Next, the algorithm checks whether $\sigma^{LLM}$ has the same step length and data format as the sampled trajectory segments. 
Once the generated trajectory passes the format check, it is converted into a state-based trajectory $\sigma^{LLM}$ and the imagined pair triple $(\sigma^{LLM}, \sigma^{0/1}, y=0)$ (i.e., self-augmented feedback) is stored into $\mathcal{D}$.

\textbf{Remark 2.} 
\textit{
The imagined trajectories do not have to adhere to physical constraints, as they serve solely for preference comparisons to train the reward model, rather than for extracting policies or constructing a world model.
We assume the reward model is Markovian, implying that the immediate reward relies exclusively on the current state.
We find that once the LLM understands the task goal, the generated trajectories, even if not physically feasible, are still beneficial for training the reward model efficiently.
This insight enables us to use a lightweight LLM with simple prompts for trajectory generation without incorporating numerous executable trajectory constraints.
}

After several rounds of querying and generating, $\mathcal{D}$ contains trajectory pairs sampled from the replay buffer $\mathcal{B}$ with LLM-judged labels, and self-augmented trajectory pairs and labels generated by the LLM. 
These labeled trajectory pairs will be used to train the reward model by minimizing Equation \ref{eqt:pref_ce}.

\section{Experiments}
\label{sec6}

\subsection{Setup}

We evaluate RL-SaLLM-F on multiple robotic manipulation tasks in the MetaWorld \cite{yu2020meta} benchmark.
The states include the coordinates of the Tool Center Point (TCP), the extent of TCP's opening, the coordinates of the manipulated object and the target position.
All of these states can be easily converted into text-based string representations.
We conduct experiments on eight tasks:
\begin{itemize}[leftmargin=2em, label=-]
    \item \textit{\textbf{Button Press}}: \ Press a button;
    \item \textit{\textbf{Drawer Open}}: \ Open a drawer;
    \item \textit{\textbf{Drawer Close}}: \ Push and close a drawer;
    \item \textit{\textbf{Door Open}}: \ Open a door with a revolving joint;
    \item \textit{\textbf{Door Unlock}}:\  Rotate the lock and unlock the door;
    \item \textit{\textbf{Window Open}}: \ Push and open a window;
    \item \textit{\textbf{Handle Pull}}: \ Pull a handle up;
    \item \textit{\textbf{Reach}}: \  Reach a goal position.
\end{itemize}

To be deemed successful, the agent must achieve the task target within a limited number of steps and maintain it until the end of the episode. 
All the positions are randomized at each initialization.
In the following sections, we aim to address five key questions:
\begin{enumerate}[leftmargin=*]
    \item Can RL-SaLLM-F master robot control without any privileged predefined rewards or human feedback? (Section \ref{sec6.2})
    \item Does each component of RL-SaLLM-F contribute to performance improvement? (Section \ref{sec6.3})
    \item Does the learned reward function align effectively with task progress? (Section \ref{sec6.4})
    \item How accurately does RL-SaLLM-F assess the quality of trajectories, and how high is the quality of the LLM-based generated trajectories? (Section \ref{sec6.5} and \ref{sec6.6})
    \item How do larger LLM impact performance? (Section \ref{sec6.7})
\end{enumerate}

\subsection{Comprehensive Performance Comparison}
\label{sec6.2}

To evaluate whether RL-SaLLM-F can master robotic manipulation tasks without relying on prior reward information, we compare it against PEBBLE\cite{pebble21} and SAC.
PEBBLE serves as a popular approach for evaluation in online PbRL and relies on the "scripted teacher" feedback based on task rewards. 
SAC directly leverages task rewards for policy learning, {representing the refer upper bound of online RL performance.}

\textbf{Remark 3.} 
\textit{
At the outset, it should be clarified that our goal is not to surpass these methods, as they both rely on privileged information from the environment, whereas RL-SaLLM-F operates without such assumptions.
Due to the lack of baseline comparisons under the same settings, we consider the comparison in this section as reference only.
}

All methods share the same hyperparameters, including the reward model structure, query number and feedback segment length.
The total query budget for the entire training process is 2000, with each session of concentrated queries consisting of 20 queries, each involving a trajectory segment length of 10.
More hyperparameters can be found in Appendix \ref{app:a}. 
Due to the cost and time constraints of queries, RL-SaLLM-F uses GPT-4o-mini-2024-07-18 as the LLM feedback. 
Each task is trained with 5e5 environment steps,and the learning curves are shown in Figure \ref{fig:exp_main}.

Overall, RL-SaLLM-F achieve comparable or even superior performance compared to PEBBLE.
Notably, in the \textit{Button Press} and \textit{Drawer Open} tasks, the success rate of RL-SaLLM-F is on par with that of SAC with real-time task rewards.
This indicates that SaLLM effectively implement feedback and leverage improved imagine trajectories to drive more efficient and superior reward function training.
Specifically, in the Button Press task, RL-SaLLM-F generated significantly higher-quality imagined trajectories (see Section \ref{sec6.6} for detail analysis).
Meanwhile, we observe that imperfect LLM feedback labels led to RL-SaLLM-F exhibiting lower training efficiency on some tasks compared to PEBBLE with "scripted teacher" feedback,
see Section \ref{sec6.5} for further analysis on label accuracy.

\subsection{Ablation Study}
\label{sec6.3}

To investigate the role of each component in RL-SaLLM-F, we conduct ablation studies. 
Specifically, we remove different modules from the algorithm the training curves of 4 chosen tasks are shown in the Figure \ref{fig:abl}. 
In the legend, 'w/o double-check' indicates executing sampling feedback and self-augmented feedback without checking the feedback labels; 
'w/o LLM feedback' indicates executing only LLM self-augmented feedback without sampling feedback; 
and 'w/o self-augmented' indicates executing only LLM sampling feedback (with double-check) without self-augmented feedback.

As Figure \ref{fig:abl} shows, when any component of RL-SaLLM-F is removed, the performance decreases.
Specifically, the absence of self-augmented feedback leads to notably poor success rate. 
On the other hand, only self-augmented feedback may lead to rapid performance gains in early training but often causes instability later on. 
The double-check mechanism serves as a valuable enhancement to the algorithm.
When incorporating self-augmented feedback, the double-check mechanism further stabilizes performance by increasing the accuracy of sampled trajectory labels.

\begin{table}[b!]
\vspace{-0.3cm}
\centering
\caption{Label accuracy and discard rate comparison for RL-SaLLM-F and the ablation variants.}
\resizebox{\linewidth}{!}{
\begin{tabular}{lcccc}
\toprule
 & \small{\textit{Button Press}} & \small{\textit{Door Open}} & \small{\textit{Door Unlock}} & \small{\textit{Handle Pull}} \\ 
\midrule
\small{\textbf{RL-SaLLM-F}}            & \textbf{72.30\%} & \textbf{65.21\%} & \textbf{71.66\%} & \textbf{67.73\%} \\
\small{w/o double-check}       & 63.06\% & 60.07\% & 66.29\% & 61.13\%  \\
\small{w/o LLM feedback}       & 62.77\% & 59.62\% & 59.64\% & 57.65\% \\
\small{w/o self-augmented}     & 64.79\% & 60.60\% & 66.25\% & 65.89\% \\
\midrule
\small{Discard Rate}               & 36.70\% & 37.89\% & 34.14\% & 37.42\% \\
\bottomrule
\end{tabular}
}
\label{tab:compare_acc}
\end{table}
\subsection{Analysis of The Learned Reward Model }
\label{sec6.4}

\begin{figure}[t]
  \centering
  \includegraphics[width=\linewidth]{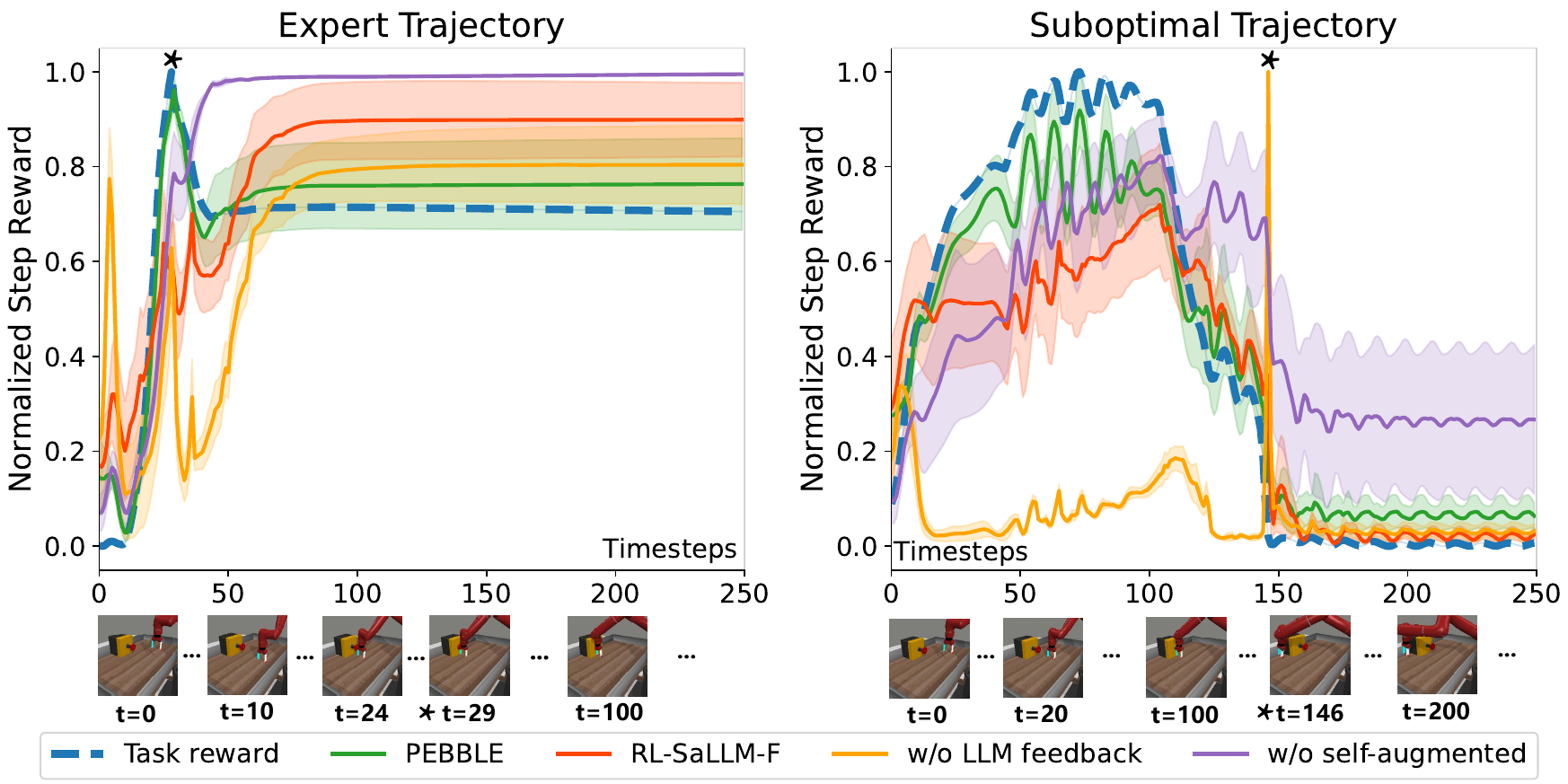}
  \caption{Normalized step rewards for the expert and suboptimal trajectories in the \textit{Button Press} task. 
  The rewards of RL-SaLLM-F show better alignment with the predefined task reward than the ablation variants, particularly in the suboptimal trajectory.
  }
  \label{fig:exp_reward}
  \vspace{-0.3cm}
\end{figure}

In addition to evaluating the quality of the learned policy by comparing the performance, 
a remaining problem is, whether the learned reward model aligns with task progress?
To answer this question, we compare the reward models learned by PEBBLE and RL-SaLLM-F with different ablation versions. 
Specifically, we choose an expert trajectory and a suboptimal trajectory in the \textit{Button Press} task, then label the rewards for these two trajectories. 
In the expert trajectory, the robotic arm moves directly to the button, presses it, and then stays still. 
In the suboptimal trajectory, the robotic arm moves for a while, presses the button, and then moves away from the button center, causing it to spring back.
The normalized step rewards of each method are shown in the Figure \ref{fig:exp_reward}.

We observe that the rewards of PEBBLE exhibit a very similar trend to the predefined task reward, which aligns with our intuition, as PEBBLE's reward model is trained relying on the task reward indirectly. 
The rewards of RL-SaLLM-F show changes that are more aligned with the task reward compared to the two ablation variants, particularly in the suboptimal trajectory.
The 'w/o LLM feedback' variant unexpectedly shows a reward peak at $t=146$, while the 'w/o self-augmented' variant maintains a high reward value from $t=130$ to $t=145$. 
This anomaly may be due to insufficient fine-grained understanding by the LLM, leading to distort generated trajectories and labeling errors.

\textbf{Remark 4.} 
\textit{
Interestingly, although the rewards of RL-SaLLM-F appear to have a larger discrepancy from the predefined rewards compared to PEBBLE, RL-SaLLM-F achieves higher task success rates. 
We suspect this may be due to the predefined reward being less effective than the goal-based evaluation reward, or the trajectory augmentation in RL-SaLLM-F contributing to greater stability in reward model training, leading to better policy improvement.
}

\begin{figure}[t]
  \centering
  \includegraphics[width=\linewidth]{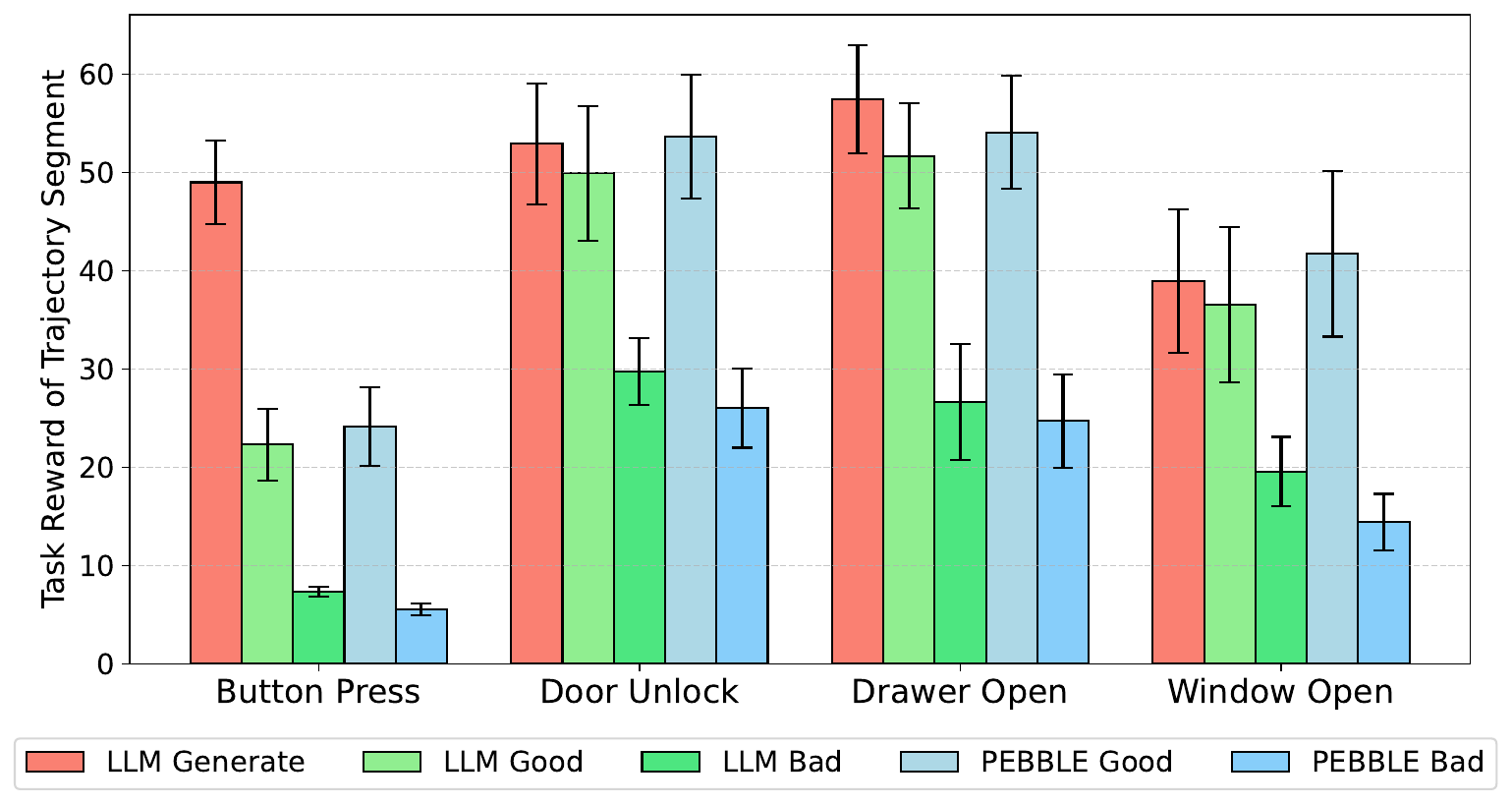}
  \caption{
  Comparison of task rewards for trajectories generated by the RL-SaLLM-F and evaluated by LLM and "scripted teacher".
  The red bars represent trajectories generated by the LLM. 
  The others represent trajectories sampled from the environment during training, with the green bars judged by the LLM and the blue bars evaluated by the "scripted teacher".
  }
  \label{fig:bar_generate}
  \vspace{-0.3cm}
\end{figure}

\subsection{Feedback Labels Quality Evaluation}
\label{sec6.5}

We further compare the quality of feedback labels to investigate the performance of our method in labeling. 
We extract all sampled trajectory pairs from the training process and analyze the accuracy of the LLM in judging these trajectory pairs (compared to the "scripted teacher"). 
The average results of five training seeds are presented in Table \ref{tab:compare_acc}. 
Apart from the header, the first four rows show the query label accuracy for RL-SaLLM-F and its ablation variants, while the last row indicates the proportion of query trajectory pairs discarded due to the the double-check mechanism in RL-SaLLM-F. 
For the 'w/o LLM feedback' variant, we still query trajectory preferences during training but does not perform  self-checking of the LLM labels. 

As shown in the Table \ref{tab:compare_acc}, RL-SaLLM-F achieve the highest label accuracy among all variants. 
The absence of the double-check mechanism lead to a decrease in accuracy, indicating that self-checking can reduce label randomness. 
Furthermore, compared to the 'w/o self-augmented' variant, the label accuracy of RL-SaLLM-F is still higher, suggesting that self-augmentation improves reward and policy performance, leading to greater diversity in sampled trajectories. 
The diversity of trajectories helps the LLM make more accurate preference judgments, consistent with our intuition in Section \ref{sec4}.
However, even the best RL-SaLLM-F method achieves only around 70\% label accuracy. 
This could be due to the bias of the "scripted teacher", which provides binary reward labels based solely on absolute reward comparisons. 
Such labels may not align with human or LLM intentions for trajectories with similar performance. Another reason is the limitations of the LLM itself. 
In Section \ref{sec6.7}, we find that the label accuracy can be further improved by a LLM with larger scale of parameters.

\subsection{Generated Trajectories Quality Evaluation}
\label{sec6.6}

\begin{figure}[t]
  \centering
  \includegraphics[width=\linewidth]{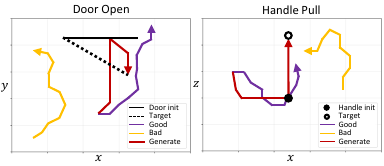}
  \caption{Two examples of 2D projections of trajectories generated by the LLM.
  Compared to the sampled trajectories, the generated trajectories successfully achieve the task goals.
  }
  \label{fig:trajectoy_generate}
  \vspace{-0.3cm}
\end{figure}

Furthermore, we study the quality of the trajectories generated by the LLM.
As analyzed in the previous section, we extract the better and worse trajectories from the pairs buffer during RL-SaLLM-F training, along with the imagined trajectories generated by the LLM, and calculate the average rewards and variance across different seeds using the predefined task reward function. 
Then, we evaluate the same trajectories using the labels of "scripted teacher" and record the average rewards, as shown in Figure \ref{fig:trajectoy_generate}. 
Only tasks related to the reward function and state are considered, as the generated trajectories do not include action information.

Comparing the rewards of the LLM-generated trajectories with those it judged as better, we find that RL-SaLLM-F can generate higher-quality trajectories than those it initially preferred, which in turn benefits reward learning.
When comparing the LLM-generated trajectories with those evaluated by the "scripted teacher", we find that the LLM-generated trajectories perform comparably to, or even better than, those deemed preferred by the ”scripted teacher“.
Moreover, a positive correlation is observed between the quality of generated trajectories and task performance, particularly in the \textit{Button Press} and \textit{Drawer Open} tasks, where RL-SaLLM-F outperforms PEBBLE. 
The average reward of the generated trajectories is also higher than that of the trajectories deemed superior by PEBBLE.

Finally, we provide two visualization examples of 2D projections of trajectories generated by the LLM in Figure \ref{fig:trajectoy_generate}.
{
In the \textit{Door Open} task, compared to the sampled trajectories, the generated one first moves directly towards the door and then pulls it open to the target position. }
In the \textit{Handle Pull} task, the generated trajectory first locates the handle and then pulls it upward to the target position. 
The examples demonstrate a thorough understanding of task goals and trajectory information by the LLM.

\subsection{Impact of LLM Scale}
\label{sec6.7}
\begin{figure}[t]
  \centering
  \includegraphics[width=\linewidth]{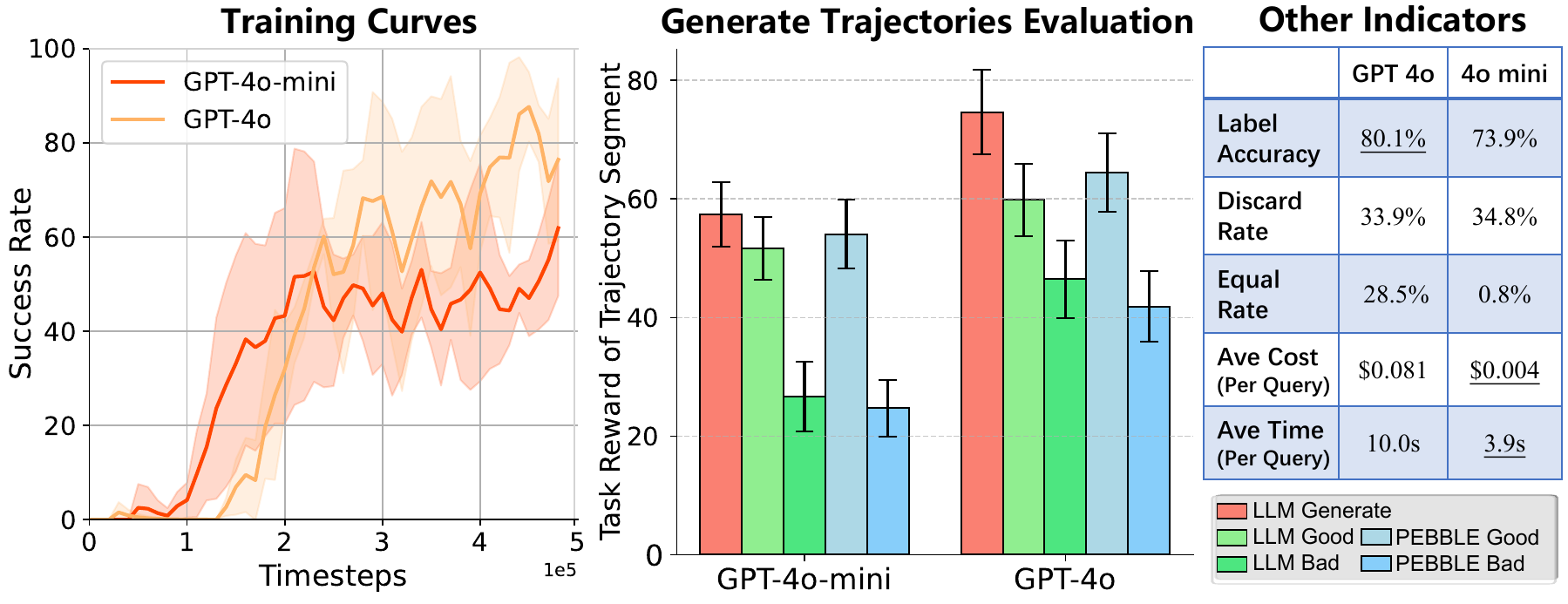}
  \caption{Comparison of RL-SaLLM-F performance with GPT-4o and GPT-4o-mini. 
  Albeit the larger LLM delivers superior results, it comes with increased resource consumption.
  }
  \label{fig:plot_vlm_compare}
\end{figure}

An intuitive question is: would scaling up the LLM boost the performance of RL-SaLLM-F? 
The answer is certainly yes.
To clearly showcase this improvement, we repeat the experimental analysis from the previous sections on the \textit{Drawer Open} task  with GPT-4o as the LLM, and conduct a comprehensive performance evaluation, as shown in Figure \ref{fig:plot_vlm_compare}. 
In addition to the familiar metrics, we define an additional one:
"Equal Rate", which is the proportion of queries where the LLM assigns equal preferences to two trajectories.

In terms of policy performance, trajectory quality, and preference label accuracy, GPT-4o significantly outperforms its smaller counterpart, GPT-4o-mini. 
Additionally, GPT-4o tends to provide more labels with $y=0.5$, suggesting a more cautious evaluation, and avoiding making arbitrary judgments when the trajectories are similar.
However, querying GPT-4o is expensive: about 20 times the price of GPT-4o-mini for the same number of tokens. 
Therefore, we still believe that combining more affordable LLMs offers a highly cost-effective solution.

\section{Conclusions}
In this work, we introduce RL-SaLLM-F, a novel technique that leverages self-augmented feedback from LLMs for online PbRL. 
By abstracting state trajectories into textual descriptions and utilizing LLMs to generate self-augmented imagined trajectories and provide preference labels, RL-SaLLM-F successfully addresses the limitations of relying on online privileged rewards or real-time human feedback. 
The experimental analysis from various perspectives have demonstrated the effectiveness of the proposed method.

Additionally, we find that the accuracy of the LLM labels in trajectory evaluation remains limited, and further improving this accuracy is essential for more efficient task training. 
Meanwhile, our method may not directly handle image inputs, future work could explore obtaining precise object coordinates through methods such as camera coordinate calibration to align with our algorithmic framework, or applying VLMs and diffusion models for discrimination and generation of image trajectories.

\section*{Acknowledgements} 
This work is supported by the National Key Research and Development Program of China under Grants 2022YFA1004000, the National Natural Science Foundation of China under Grants 62173325, and the CAS for Grand Challenges under Grants 104GJHZ2022013GC.

\clearpage
\bibliographystyle{ACM-Reference-Format} 
\bibliography{sample}


\clearpage
\onecolumn
\appendix

\begin{center}
    \LARGE \textbf{Appendix}
\end{center}

\section{Experimental Details}
\label{app:a}
\subsection{Training details}
We trained all methods using the full set of hyperparameters listed in Table \ref{tab:hyperparameter}, and the agents are pre-trained for 10,000 timesteps.

\begin{table}[htbp]
\centering
\caption{Hyperparameter values used in the experiments.}
\begin{tabular}{l l}
\toprule
\textbf{Hyperparameter}            & \textbf{Value}                            \\ \midrule
Training steps                     & 500,000  \\
Initial temperature                & 0.1                                      \\
Segment of length                  & 10            \\
Learning rate                      & 0.0003  \\
Critic target update frequency     & 2                                        \\
$(\beta_1, \beta_2)$               & (0.9, 0.999)                             \\
Frequency of feedback              & 5000   \\
Hidden units per each layer        & 256        \\
Number of layers                   & 3            \\
Batch size                         & 512        \\
Optimizer                          & Adam                                 \\
Critic EMA $\tau$                  & 0.005                                     \\
Discount $\gamma$                  & 0.99                                      \\
Maximum budget                     & 2,000    \\
Number of queries per session      & 20 \\ \bottomrule
\label{tab:hyperparameter}
\end{tabular}

\end{table}

\subsection{Reward model}
For the reward model, we use a three-layer neural network with 256 hidden units in each layer and leaky ReLU for activation.
To enhance the stability of reward learning, we use an ensemble of three reward models whose outputs are constrained by a tanh function.
Each model is trained by optimizing the cross entropy loss defined in Equation \ref{eqt:pref_ce} using the Adam optimizer with an initial learning rate of 0.0003. 
The reward models are trained for 10 epochs with each update.

\subsection{Scripted Teacher}
As online human preference is always not available, the "scripted teacher" is used to provide
preference labels $y$ in most previous works. 
Specifically, a deterministic teacher, which generates preferences based on predefined task reward $r$ from the environment as follows: 
$$y=i, \ \ \text{where} \ \ i=\arg \max_{i \in \{0,1\}} \sum_{t} r(s_t^i,a_t^i)$$.

\clearpage
\section{Additional Experimental Results}
\subsection{Environment Setup}
\begin{figure}[h]
  \centering
  \includegraphics[width=\linewidth]{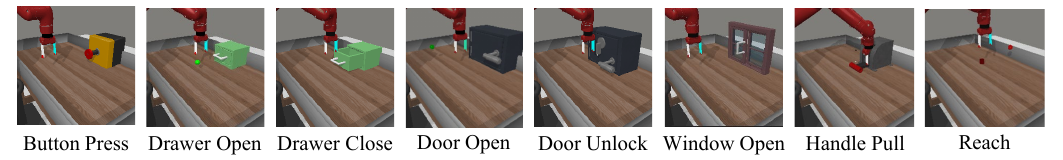}
  \caption{Illustration of partial tasks in  MetaWorld.}
  \label{fig:env}
\end{figure}

We provide the visual illustration of tasks in the experiment in Figure \ref{fig:env}.

Additionally, each environment has a fixed 37-dimensional state space, some of which involve object rotations that are not useful for our tasks. 
For all algorithms (including RL-SaLLM-F, PEBBLE, and SAC), we apply appropriate dimension reduction. 
Specifically, we retain only the coordinates of the Tool Center Point (TCP), the extent of the TCP’s opening, the coordinates of the manipulated object, and the target position. 
These coordinates are kept for both the current and previous time steps in each state. 
Since the target position is typically fixed during the episode, we retained only one target value. 
As a result, for the first seven tasks, the state is reduced from 39 dimensions to 17 dimensions ($(3+1+3)\times2+3$). 
For the last \textit{Reach} task, we further remove the coordinates of the manipulated object and the extent of the TCP’s opening, reducing the state to 9 dimensions ($3\times2+3$).

\subsection{Training Curves in Predefined Task Return}

\begin{figure}[h]
  \centering
  \includegraphics[width=\linewidth]{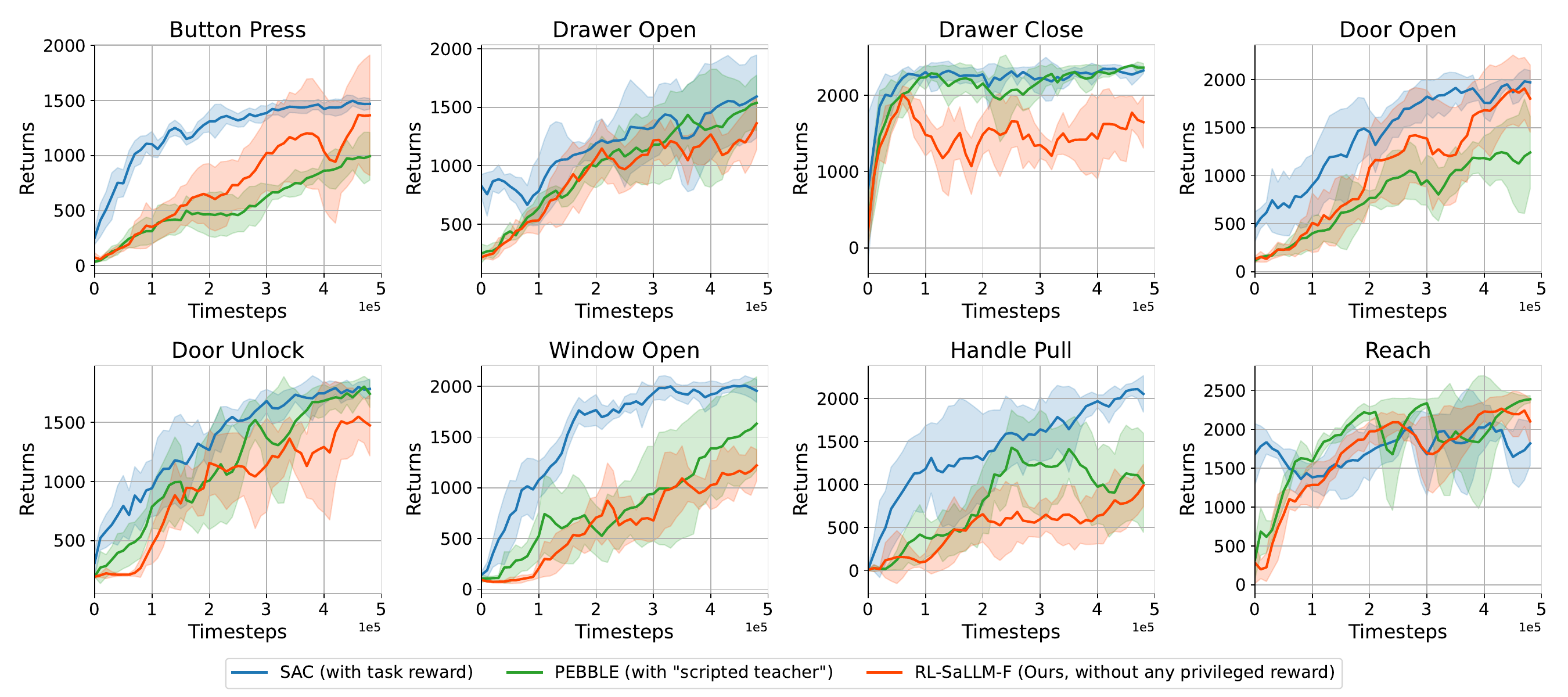}
  \caption{Learning curves of all compared methods on 8 tasks, the vertical axis represents the trajectory return.}
  \label{fig:exp_main_reward}
\end{figure}

\begin{figure}[h]
  \centering
  \includegraphics[width=\linewidth]{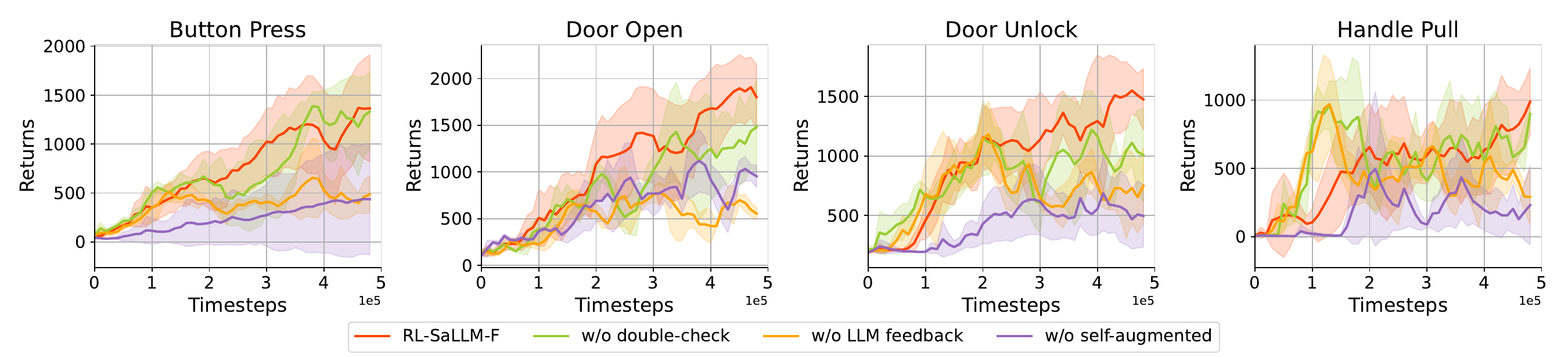}
  \caption{Learning curves of the ablation study, the vertical axis represents the trajectory returns.}
  \label{fig:exp_ablation_reward}
\end{figure}

In addition to reporting task success rates, we also record trajectory return curves calculated using predefined rewards during training in Figure \ref{fig:exp_main_reward} and Figure \ref{fig:exp_ablation_reward}. Although these two metrics differ, the final results and trends are largely consistent with the figures presented in the main text. This indicates that RL-SaLLM-F not only successfully completes tasks but also maintains high-quality performance throughout the process.

\subsection{Comparison with VLM Feedback Methods}
\label{sec6.8}

Moreover, we compare RL-SaLLM-F with RL-VLM-F \cite{wang2024rl}, which uses GPT-4 for querying but generates preference feedback through pairwise image comparisons.
Admittedly, RL-VLM-F learns a reward model from image observations, which may be biased compared to our approach that learns from state observations. 
Thus, we consider the comparison results as merely a reference. 
We find that RL-VLM-F nearly failed in the \textit{Drawer Open} task, likely due to the limited dynamic information provided by feedback from a single pair of images compared to trajectories. 
Due to space constraints, detailed results are recorded in Appendix \ref{app:b3}.
In addition, there is a noteworthy method, RL-HT-F, which requires real-time human feedback and extracts key states from human textual feedback for reward learning \cite{urcelay2024reinforcement}. 
We qualitatively compare RL-SaLLM-F with these two methods in Table \ref{tab:compare_method}, to visually showcase the inherent characteristics and advantages of RL-SaLLM-F.

\begin{table}[h]
\centering
\caption{Qualitative comparison of different methods.}
\begin{tabular}{llll}
\toprule
\textbf{}                      & \textbf{RL-SaLLM-F}           & \textbf{RL-VLM-F \cite{wang2024rl}}           & \textbf{RL-HT-F \cite{urcelay2024reinforcement}}             \\ 
\midrule
\textbf{Human-in-the-loop}      & No                            & No                          & Yes                          \\ 
\textbf{LLM Role}               & Judge \& Generate             & Judge                       & Extract Human Text           \\ 
\textbf{LLM Judging Type}       & Trajectories                  & Images                      & \textbackslash{}             \\ 
\textbf{LLM Judging Length}     & Long Trajectories             & Just one state              & \textbackslash{}             \\ 
\textbf{LLM Types}              & Text \& Multimodal            & Multimodal                  & Text \& Multimodal           \\
\textbf{Price}      & Depend on the LLM       & High (Multimodal input)                      & Depend on the LLM      \\ 
\textbf{Training Efficiency}    & {High}                          & Medium (Image feedback)   & Low (Human-in-the-loop)      \\
\textbf{Applicable Scope}       & State in text format  & Markov reward               & Human give text feedback   \\ 
\bottomrule
\end{tabular}
\label{tab:compare_method}
\end{table}

\subsection{More Comparison with RL-VLM-F}
\label{app:b3}
\begin{figure*}[h]
\centering
\subfloat[Comparison of Training Curves.]{
\includegraphics[width=.53\linewidth]{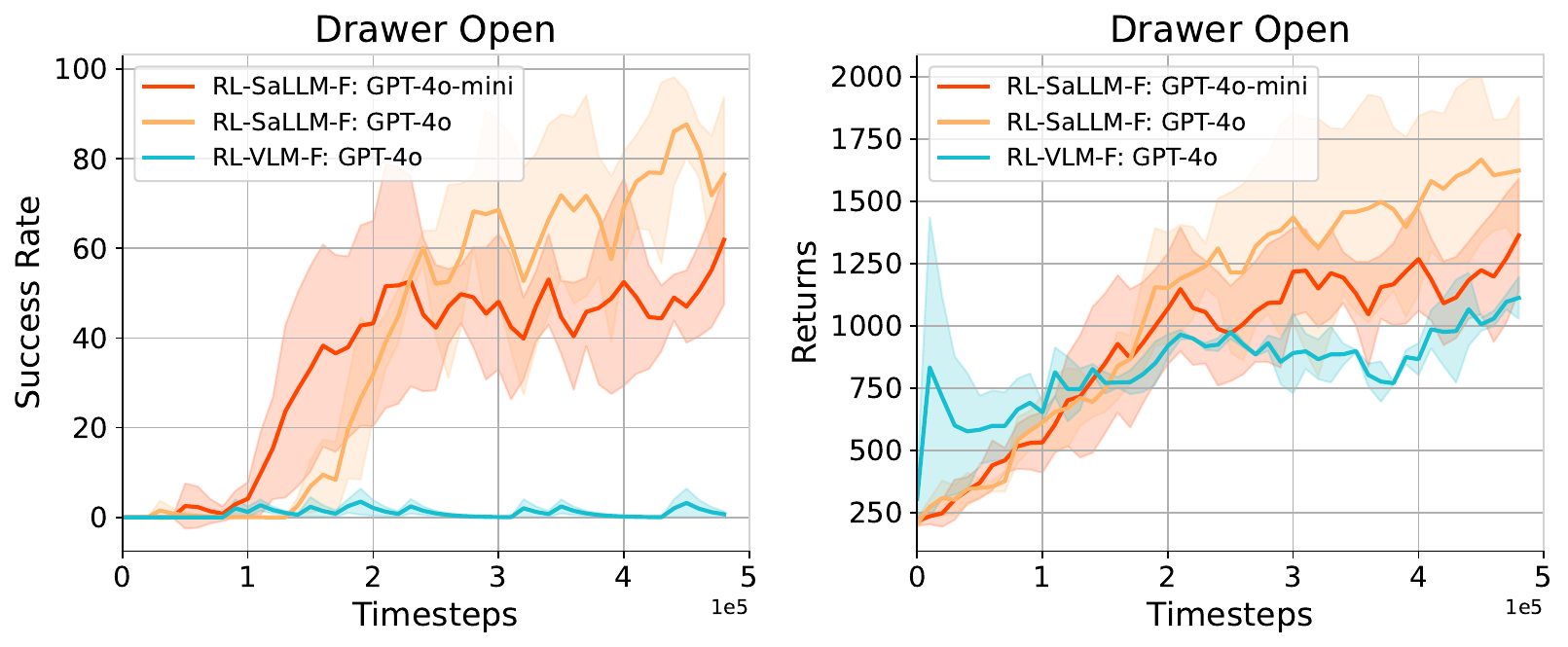}
\label{fig:plot_rlvlmf_compare1}
}
\subfloat[Comparison of Typical Policies.]{
\includegraphics[width=.45\linewidth]{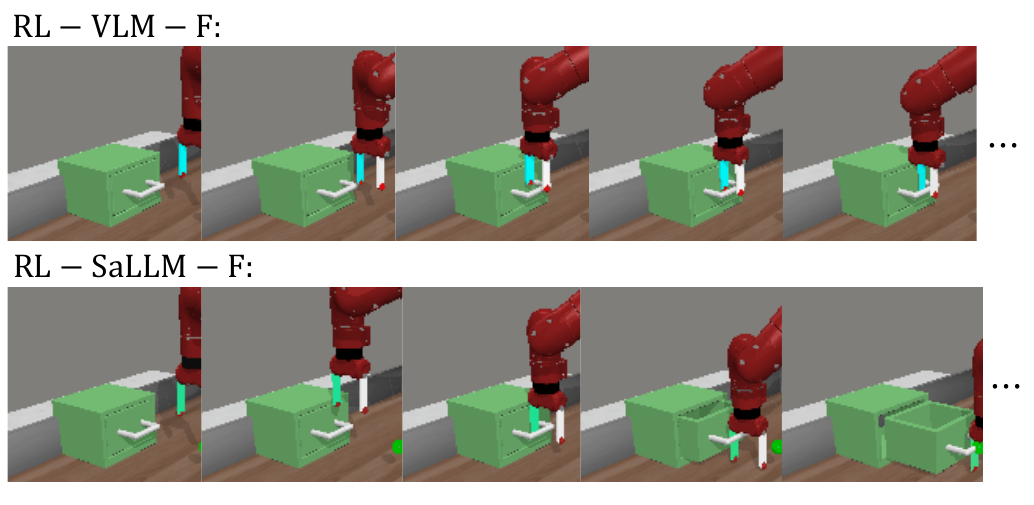}
\label{fig:plot_rlvlmf_compare2}
}
\caption{
    Comparison of RL-VLM-F and RL-SaLLM-F. 
}
\label{fig:plot_rlvlmf_compare}
\end{figure*}

We further compare RL-SaLLM-F with RL-VLM-F, using GPT-4o as the VLM to provide preference feedback. 
Due to the high cost of GPT-4o queries, we conducte experiments only on the \textit{Drawer Open} task. 
It is worth noting that the original RL-VLM-F paper \cite{wang2024rl} also conducted experiments on the same task. 
Here, while reproducing RL-VLM-F, we maintain the same hyperparameters as RL-SaLLM-F, but there are two key setting differences from the original paper \cite{wang2024rl}:
\begin{enumerate}
    \item The environment in  \cite{wang2024rl} is deterministic \footnote{\href{https://github.com/yufeiwang63/RL-VLM-F/issues/3\#issuecomment-2278443301}{https://github.com/yufeiwang63/RL-VLM-F/issues/3\#issuecomment-2278443301}}, meaning the TCP and target positions are the fixed at each initialization, whereas we use random initialization;
    \item \cite{wang2024rl} processed the input by removing the robotic arm from the images.
    We believe this approach may not adequately model the movement information of the robot in the reward function, so we retain a third-person observational perspective as the input of VLM and reward model, as shown in the Figure \ref{fig:plot_rlvlmf_compare2}.
\end{enumerate}

Figure \label{fig:plot_rlvlmf_compare} compares the performance of RL-VLM-F and RL-SaLLM-F, highlighting training curves and typical policies.
The training curves in Figure \ref{fig:plot_rlvlmf_compare1} show that RL-SaLLM-F (both GPT-4o-mini and GPT-4o) significantly outperforms RL-VLM-F (GPT-4o). 
RL-VLM-F almost fails to complete the task, but due to the task rewards being tied to both the drawer's open state and the proximity of the robotic arm to the drawer, the trajectory returns remain fairly decent.
Upon further examination of its typical policy trajectories, we find that while the policy of RL-VLM-F accurately locates the drawer handle, it fails to pull it open, likely due to the lack of self-augmented exploration. 
This lack potentially hinders the reward model from learning important nuances, ultimately leading to suboptimal policy performance.
In contrast, RL-SaLLM-F not only locates the handle but successfully pulls the drawer open, demonstrating more effective task execution.

It is important to emphasize that this comparison might be somewhat biased, as RL-VLM-F learns the reward function from images. Therefore, the results of the comparison should be considered as a reference only.

\subsection{Comparison with Reward-free Baselines}

Since no direct baseline uses text descriptions, we compare with popular reward-free methods that learn rewards from observations using pretrained models. As suggested, we directly compare the results presented in \cite{luo2024text} on six tasks. 
All three baselines leverage image or video models to assist in reward design. Similar to our setting, they do not rely on any environmental feedback but instead learn rewards from observations to support downstream RL training.
Specifically, RoboClip \cite{sontakke2024roboclip} uses pretrained VLMs to generate rewards from a single demonstration for imitation learning in robotics; 
VLM-RM \cite{rocamonde2024vision} leverages pretrained vision-language models, like CLIP, as zero-shot reward models for reinforcement learning, enabling agents to learn tasks directly from natural language descriptions; 
TADPoLe \cite{luo2024text} utilizes pretrained, frozen text-conditioned diffusion models to generate dense, zero-shot reward signals for training RL policies.

We provide a detailed comparison of success rate metrics for each task in Table \ref{tab:performance_rf_baseline}. Due to the robustness of the text descriptions and our self-checking and self-enhancing PbRL mechanism, our method outperforms other baselines.

\begin{table}[h]
\centering
\caption{Performance comparison with reward-free baselines.}
\begin{tabular}{lcccc}
\toprule
Task & RoboClip & VLM-RM & TADPoLe & RL-SaLLM-F (Ours) \\ 
\midrule
Door Open & 0 (±0) & 0 (±0) & 40.0 (±49.0) & \textbf{89.5} (±7.6) \\ 
Drawer Open & 0 (±0) & 10.0 (±30.0) & 45.3 (±46.6) & \textbf{61.9} (±14.4) \\ 
Drawer Close & 44.5 (±4.6) & \textbf{100.0} (±0) & \textbf{100.0} (±0) & 74.1 (±12.3) \\ 
Window Open & 0 (±0) & 9.7 (±29.0) & 74.0 (±37.9) & \textbf{84.5} (±13.9) \\ 
Button Press & 0 (±0) & 30.0 (±45.8) & 73.0 (±38.5) & \textbf{83.1} (±19.8) \\ 
Handle Pull & 0 (±0) & 11.3 (±18.3) & 0 (±0) & 57.1 (±5.72) \\ 
\midrule
Average & 7.41 & 26.83 & 55.38 & \textbf{75.03} \\ 
\bottomrule
\end{tabular}
\label{tab:performance_rf_baseline}
\end{table}

\subsection{Point-Maze Benchmark Evaluation}

We add results in the point-maze environment \cite{brockman2016openai} with 3 point-ball navigation tasks. 
For intuitive representation, we directly show the average test score of the last checkpoint. The detailed results are in Table 1.
SAC performs poorly in the Point-Maze environment, likely because the predefined reward is solely based on the Euclidean distance between the current position and the target, which may not be reliable in a maze environment.
RL-SaLLM-F outperforms PEBBLE, showing the strong generalization ability LLM brings to our approach across domains.

\begin{table}[h]
\centering
\caption{Performance comparison in Point-Maze benchmark.}
\begin{tabular}{lcccc}
\toprule
Task & SAC (dense reward) & SAC (sparse reward) & PEBBLE & RL-SaLLM-F \\ 
\midrule
PointMaze\_Open & 37.3 (±7.6) &  30.0 (±9.5)  & \textbf{58.7} (±5.8) & 49.9 (±4.3) \\ 
PointMaze\_UMaze &  44.3 (±7.7)  &  56.0 (±9.2)  & 45.0 (±10.9) & \textbf{56.1} (±8.5) \\ 
PointMaze\_Medium &  5.6 (±2.5)  &  5.6 (±3.3)  & 6.7 (±3.5) & \textbf{11.8} (±3.3) \\ 
\midrule
Average & 29.07 & 30.5 & 36.8 & \textbf{39.3} \\ 
\bottomrule
\end{tabular}
\label{tab:comparison}
\end{table}

\subsection{Comparison With DT-based Generated Trajectory}

\begin{figure}[h]
  \centering
  \includegraphics[width=.7\linewidth]{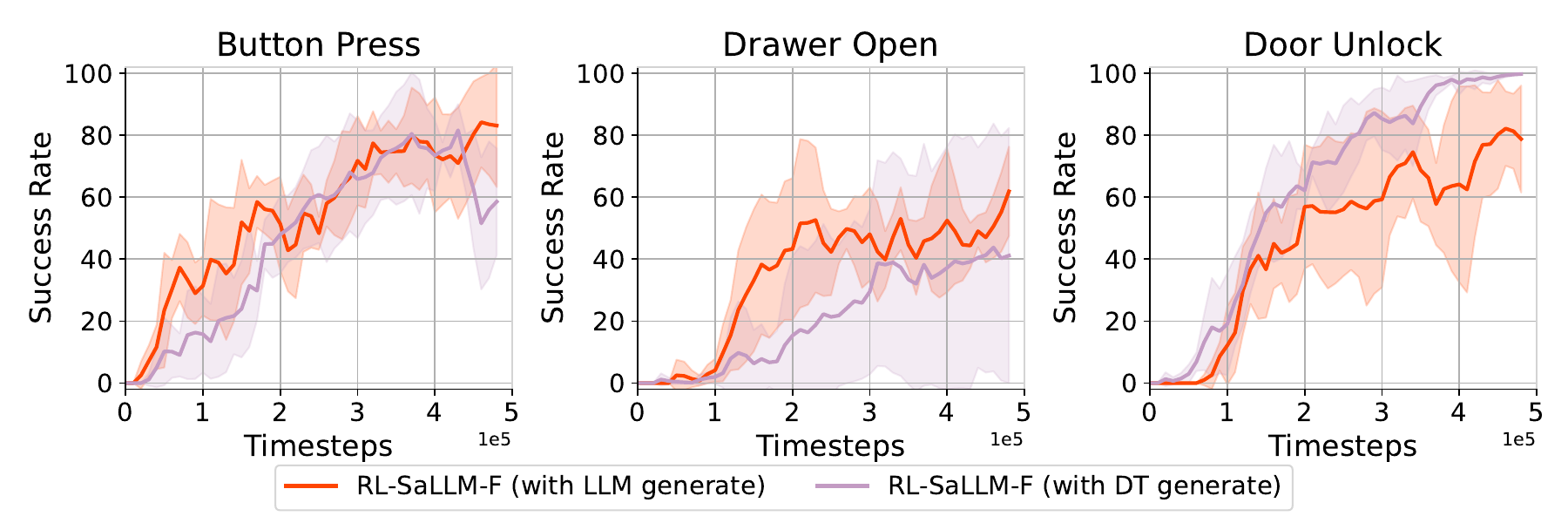}
  \caption{Learning curves of LLM and DT as trajectory generator. }
  \label{fig:llm_vs_dt}
\end{figure}

Decision Transformer (DT) \cite{chen2021decision} relies on return-to-go alongside state-action representations. For reward-free offline data, DT requires expert data. 
To compare with DT, we first train DT on 30 prior expert trajectories, then use it to generate "imaginary trajectories" instead of LLM.
The detailed results of the success rate training curves for three tasks are given in Figure \ref{fig:llm_vs_dt}.

As the number of expert trajectories increases, the DT method will yield better performance.
However, in the field of robotic operations, expert trajectories are difficult to obtain.
Additionally, DT trained on offline data suffers from severe out-of-distribution (OOD) issues, often performing worse than trajectories generated by LLMs. 
RL-SaLLM-F does not rely on prior expert trajectories or return-to-go information, which just requires online reward-free trajectories in the experience buffer.

\subsection{Evaluating Reward Model Bias With Public Offline Dataset}
We present new experimental results using public offline PbRL dataset\footnote{\href{https://drive.google.com/file/d/1lo5Wt9Go_E_5c8ymfXFvsTY6kDUelqu2/view}{https://drive.google.com/file/d/1lo5Wt9Go\_E\_5c8ymfXFvsTY6kDUelqu2/view}}, which includes mixed data from an agent with a 50\% success rate, to evaluate reward model bias \cite{cpl24}. The label accuracy of various reward models on three offline datasets is shown in Table \ref{tab:acc_pub}.

RL-SaLLM-F has more training coverage of successful trajectories, so it is easier to distinguish good and bad trajectories in offline data of mixed quality; the accuracy difference with the variant w/o self-augmented is even larger than that of the online evaluation.

\begin{table}[h]
\centering
\caption{Label accuracy in public offline dataset}
\begin{tabular}{lcc}
\toprule
Task & RL-SaLLM-F  & w/o self-augmented  \\ 
\midrule
Drawer Open & \textbf{71.08\%} & 64.04\% \\ 
Door Open & \textbf{74.50\%} & 69.87\% \\ 
Button Press & \textbf{77.31\%} & 66.74\% \\ 
Average & \textbf{74.53\%} & 66.88\% \\ 
\bottomrule
\end{tabular}
\label{tab:acc_pub}
\end{table}

\subsection{Resource Consumption}

We analyze token consumption for a training cycle. 
The detailed average token consumption and query costs (in USD) per training cycle for each ablation variant is listed in Table \ref{tab:token_cost_comparison}.
When double-check fails, no extra queries are made, so the token consumption difference is similar. 
Although RL-SaLLM-F has the highest consumption among the variants, it remains at an affordable and acceptable cost level.

\begin{table}[ht]
\centering
\caption{Comparison of tokens and cost across different variations of RL-SaLLM-F.}
\begin{tabular}{lccc}
\toprule
Method & Total Input Tokens & Total Output Tokens & Cost (\$) \\ 
\midrule
RL-SaLLM-F & $\approx$  6.07M & $\approx$  2.14M & $\approx$  2.19 \\ 
w/o double-check & $\approx$  5.31M & $\approx$  1.59M & $\approx$  1.75 \\ 
w/o LLM feedback & $\approx$  3.27M & $\approx$  0.74M & $\approx$  0.93 \\ 
w/o self-augmented & $\approx$  4.08M & $\approx$  1.69M & $\approx$  1.63 \\ 
\bottomrule
\end{tabular}
\label{tab:token_cost_comparison}
\end{table}

\clearpage
\section{Prompts \& Examples}
\subsection{Prompts}
\label{sec_prompt}

We provide the prompt used for querying the LLM in the \textit{Button Press} task, where {} represents the steps and trajectory information inserted by the program. 
For other tasks, the goal description is modified according to the task objective, and other details are adjusted accordingly. 
In the future, we will open-source the entire code of RL-SaLLM-F, where all the prompts used for querying in each task will be available in detail.

\begin{tcolorbox}[colback=gray!5!white, colframe=gray!75!black, title=Prompt for querying preference lables]
\begin{lstlisting}[style=mystyle]
Suppose you are a good robot trajectory evaluator. Now you need to evaluate the quality of the robot's motion trajectory.
The goal is to control the Tool Center Point (TCP) of the robot to press a button.
The following are two trajectories, which contain {} steps, where:
(1) "tcp" represents the end position of the robot actuator, which is expressed in three-dimensional Cartesian coordinates in the range of [0,1];
(2) "obj" represents the object position that the robot needs to touch, which is expressed in three-dimensional Cartesian coordinates in the range of [0,1];
(3) "target" represents the position of the target button, which is expressed in three-dimensional Cartesian coordinates in the range of [0,1];
    {}
Please answer the following two questions step by step: 
1. Is there any difference between Trajectory 1 and Trajectory 2 in terms of achieving the goal? 
Reply your analysis.
2. Which trajectory you think do better with achieving the goal?
Reply a single line of 1 if you think the goal is better achieved in Trajectory 1, or 2 if it is better achieved in Trajectory 2. Reply 0 if the text is unsure or there is no significantly difference.
\end{lstlisting}
\end{tcolorbox}

\begin{tcolorbox}[colback=gray!5!white, colframe=gray!75!black, title=Prompt for generating better trajectories]
\begin{lstlisting}[style=mystyle]
Based on your analysis, Can you generate a new trajectory based on the initial state of that good trajectory that you think can better achieve the goal?
The generated trajectory should meet the following characteristics:
(1) The movement of TCP should be smooth and touch obj as quickly as possible, then the change of obj should conform to the laws of physics, change smoothly, avoid sudden changes in coordinates and finally the obj should reach the target;
(2) TCP should first move to the position of obj, that is, the coordinates of TCP and obj should be at similar values, and then push obj to move;
(3) Output a trajectory that conforms to the input trajectory format, the step size should be {} and the trajectory should be started with
    {}
Replay only the generate better trajectory. 
\end{lstlisting}
\end{tcolorbox}

\clearpage
\subsection{Q\&A Examples}
\label{app:d}
We give an example of a real query in \textit{Button Press} task to help understand the working process of RL-SaLLM-F clearly.
Each complete query should consist of three inputs and outputs:

\begin{tcolorbox}[colback=gray!5!white, colframe=gray!75!black, title=Input 1: Label querying]
\begin{lstlisting}[style=mystyle]
Suppose you are a good robot trajectory evaluator. Now you need to evaluate the quality of the robot's motion trajectory.
The goal is to control the Tool Center Point (TCP) of the robot to press a button.
The following are two trajectories, which contain 10 steps, where:
(1) "tcp" represents the end position of the robot actuator, which is expressed in three-dimensional Cartesian coordinates in the range of [0,1];
(2) "obj" represents the object position that the robot needs to touch, which is expressed in three-dimensional Cartesian coordinates in the range of [0,1];
(3) "target" represents the position of the target button, which is expressed in three-dimensional Cartesian coordinates in the range of [0,1];

Trajectory 1:
{
    "tcp":[
        [-0.0567,0.8098,0.4486],[-0.0652,0.8094,0.4480],[-0.0780,0.8093,0.4443],[-0.0921,0.8085,0.4426],
        [-0.1055,0.8053,0.4474],[-0.1151,0.8005,0.4558],[-0.1161,0.7965,0.4629],[-0.1113,0.7944,0.4672],
        [-0.1080,0.7924,0.4709],[-0.1078,0.7896,0.4745]
    ];
    "obj":[
        [-0.0229,0.6803,0.1150],[-0.0229,0.6803,0.1150],[-0.0229,0.6803,0.1150],[-0.0229,0.6803,0.1150],
        [-0.0229,0.6803,0.1150],[-0.0229,0.6803,0.1150],[-0.0229,0.6803,0.1150],[-0.0229,0.6803,0.1150],
        [-0.0229,0.6803,0.1150],[-0.0229,0.6803,0.1150]
    ];
    "target":
        [-0.0229,0.7739,0.1150];
};
Trajectory 2:
{
    "tcp":[
        [0.4363,0.8715,0.4302],[0.4358,0.8708,0.4314],[0.4355,0.8706,0.4304],[0.4355,0.8710,0.4280],
        [0.4358,0.8719,0.4254],[0.4363,0.8730,0.4208],[0.4372,0.8744,0.4149],[0.4381,0.8760,0.4084],
        [0.4392,0.8778,0.3985],[0.4404,0.8801,0.3917]
    ],
    "obj":[
        [0.0760,0.6951,0.1150],[0.0760,0.6951,0.1150],[0.0760,0.6951,0.1150],[0.0760,0.6951,0.1150],
        [0.0760,0.6951,0.1150],[0.0760,0.6951,0.1150],[0.0760,0.6951,0.1150],[0.0760,0.6951,0.1150],
        [0.0760,0.6951,0.1150],[0.0760,0.6951,0.1150]
    ],
    "target":
        [0.0760,0.7887,0.1150];
};

Please answer the following two questions step by step: 
1. Is there any difference between Trajectory 1 and Trajectory 2 in terms of achieving the goal? 
Reply your analysis.
2. Which trajectory you think do better with achieving the goal?
Reply a single line of 1 if you think the goal is better achieved in Trajectory 1, or 2 if it is better achieved in Trajectory 2. Reply 0 if the text is unsure or there is no significantly difference.
\end{lstlisting}
\end{tcolorbox}

\begin{tcolorbox}[colback=gray!5!white, colframe=gray!75!black, title=Output 1: Label querying]
\begin{lstlisting}[style=mystyle]
To evaluate the two trajectories, we need to consider how closely the TCP positions approach the object and target button positions at each step. 
1. Differences in Achieving the Goal:
Trajectory 1:
    TCP Positions: The TCP moves from approximately [-0.0567, 0.8098, 0.4486] to [-0.1078, 0.7896, 0.4745].
    Object Position: The object is fixed at [-0.0229, 0.6803, 0.1150].
    Target Position: The target position is [-0.0229, 0.7739, 0.1150].
    In Trajectory 1, the TCP positions do not come close to the object or target positions, remaining at X-coordinates well below both the object and target positions, indicating that the robot is not effectively reaching either the button or the object it must interact with.
Trajectory 2:
    TCP Positions: The TCP moves from approximately [0.4363, 0.8715, 0.4302] to [0.4404, 0.8801, 0.3917].
    Object Position: The object remains at [0.0760, 0.6951, 0.1150].
    Target Position: The target position is [0.0760, 0.7887, 0.1150].
    In Trajectory 2, although the TCP positions primarily stay above the object and target positions, they traverse along a path closer to them as time progresses, indicating a better approach toward interacting with the object and the target.
Overall Analysis: 
    Trajectory 1 has a significant issue of not approaching either the object or target effectively; its TCP remains far from both.
    Trajectory 2 has a more consistent trajectory that gradually approaches the object and is more closely aligned with the target position, suggesting a better performance toward achieving the goal.
2. Which Trajectory is Better?
    2
\end{lstlisting}
\end{tcolorbox}

\begin{tcolorbox}[colback=gray!5!white, colframe=gray!75!black, title=Input 2: Label double-check]
\begin{lstlisting}[style=mystyle]
# Same as input1, except the order of the tracks is swapped.
......
Trajectory 1:
    {
    "tcp":[
        [0.4363,0.8715,0.4302],[0.4358,0.8708,0.4314],[0.4355,0.8706,0.4304],[0.4355,0.8710,0.4280],
    ......
    };
Trajectory 2:
    {
    "tcp":[
        [-0.0567,0.8098,0.4486],[-0.0652,0.8094,0.4480],[-0.0780,0.8093,0.4443],[-0.0921,0.8085,0.4426],
    ......
    };
......
\end{lstlisting}
\end{tcolorbox}

\begin{tcolorbox}[colback=gray!5!white, colframe=gray!75!black, title=Output 2: Label double-check]
\begin{lstlisting}[style=mystyle]
# Trajectory analysis is similar to Output2, so it is omitted.
......
2. Which Trajectory is Better?
    1
# Double-check: Pass
\end{lstlisting}
\end{tcolorbox}

\begin{tcolorbox}[colback=gray!5!white, colframe=gray!75!black, title=Input 3: Better Trajectory Generation]
\begin{lstlisting}[style=mystyle]
Based on your analysis, Can you generate a new trajectory based on the initial state of that good trajectory that you think can better achieve the goal?
The generated trajectory should meet the following characteristics:
(1) The movement of TCP should be smooth and touch obj as quickly as possible, then the change of obj should conform to the laws of physics, change smoothly, avoid sudden changes in coordinates and finally the obj should reach the target;
(2) TCP should first move to the position of obj, that is, the coordinates of TCP and obj should be at similar values, and then push obj to move;
(3) Output a trajectory that conforms to the input trajectory format, the step size should be 10 and the trajectory should be started with
{
    "tcp":[
        [0.4363,0.8715,0.4302],
    ],
    "obj":[
        [0.0760,0.6951,0.1150];
    ],
    "target":
        [0.0760,0.7887,0.1150];
}
Replay only the better generated trajectory. 
\end{lstlisting}
\end{tcolorbox}

\begin{tcolorbox}[colback=gray!5!white, colframe=gray!75!black, title=Onput 3: Better Trajectory Generation]
\begin{lstlisting}[style=mystyle]
{
    "tcp":[
        [0.4363,0.8715,0.4302], [0.4360,0.8705,0.4305], [0.4357,0.8695,0.4310], [0.4355,0.8685,0.4315], 
        [0.4352,0.8675,0.4320], [0.0755,0.7065,0.1150], [0.0750,0.7180,0.1150], [0.0745,0.7300,0.1150], 
        [0.0740,0.7400,0.1150], [0.0735,0.7887,0.1150]
    ],
    "obj":[
        [0.0760,0.6951,0.1150], [0.0760,0.6951,0.1150], [0.0760,0.6951,0.1150], [0.0760,0.6951,0.1150],
        [0.0760,0.6951,0.1150], [0.0755,0.7065,0.1150], [0.0750,0.7180,0.1150], [0.0745,0.7300,0.1150],
        [0.0740,0.7400,0.1150], [0.0735,0.7887,0.1150]
    ],
    "target":
        [0.0760,0.7887,0.1150];
};
\end{lstlisting}
\end{tcolorbox}

\subsection{One-shot Prompt Comparison}
In the original zero-shot prompt, only questions were asked without providing demonstration answers. As shown in Appendix \ref{sec_prompt}.
We conduct an experiment to test zero-shot and one-shot prompts (with the addition of a demonstration) in the button press task. In the one-shot prompt, we included a standard example of an answer as an additional reference. Detailed prompts are provided below.
The results showed that label accuracy decreased slightly from 72.3\% to 71.8\%. 
This indicates that query formulations have little impact on label accuracy.

\begin{tcolorbox}[colback=gray!5!white, colframe=gray!75!black, title=An Example of One-shot Prompt]
\begin{lstlisting}[style=mystyle]
1. Analysis:
    Trajectory 1:*
        The 'tcp' positions gradually change over the 10 steps but remain relatively close to each other and are consistent in their path.
        The 'obj' position remains constant at [0.0451,0.7300,0.0900] throughout the trajectory.
        The 'target' is at [0.0451,0.5400,0.0900], indicating that to reach the target, the TCP should be pulling the object handle slightly downward considering the target's Y-coordinate.
    Trajectory 2:
    ...
2. Result
        Based on the analysis, Trajectory 2 appears to make a more concerted effort to move towards the target's Y-coordinate, albeit with some fluctuations in the path.
    Therefore:
        *2*
\end{lstlisting}
\end{tcolorbox}


\end{document}